\documentclass[
]{ceurart}

\usepackage{listings}
\begin{document}

\copyrightyear{2026}
\copyrightclause{Copyright for this paper by its authors.
  Use permitted under Creative Commons License Attribution 4.0
  International (CC BY 4.0).}

\conference{CLEF 2026 Working Notes, 21 -- 24 September 2026, Jena, Germany}

\title{Team DACTYL at PAN 2026: Bayesian Data Mixing and Empirical X-risk Minimization for AI-text Detection}
\title[mode=sub]{Notebook for the PAN Lab at CLEF 2026}

\author[1]{Shantanu Thorat}[%
orcid=0009-0004-8470-0713,  
email=st980[at]cantab.ac.uk,
url=https://shantanuthorat.io,
]\cormark[1]

\address[1]{College Station, Texas, US}


\cortext[1]{Work completed independently of employment at Texas A\&M University.}

\begin{abstract}
 Existing research shows that AI-generated text detection classifiers achieve strong in-distribution (ID) performance but do not maintain the same performance on out-of-distribution (OOD) texts, suggesting overfitting to dataset-specific features. However, combining different training datasets doesn't always improve performance and, in some cases, can even encourage shortcut learning. To address this issue, we fine-tune BERT-tiny models with Bayesian classification heads to select texts across three different datasets to use as a consolidated training set. We trained three different classifiers: fine-tuned DeBERTa-V3-large and ModernBERT-large classifiers via empirical X-risk minimization, and an MCGrad model that calibrates the predictions from the ModernBERT-large classifier. The DeBERTa-V3-large-large classifier achieves a mean score of 0.882 on the PAN 2026 test set across five metrics: AUROC, $F_1$, C@1, Brier score, and $F_{0.5u}$. ModernBERT-large achieves a score of 0.96 while MCGrad achieves the best score of the three with a mean score of 0.974, ranking second on the leaderboard. Our results highlight that careful dataset curation can lead to strong OOD performance. We release our ModernBERT-large and DeBERTa-V3-large models at \url{https://huggingface.co/collections/ShantanuT01/panclef-2026}. 
\end{abstract}

\begin{keywords}
  Bayesian neural networks \sep
  empirical X-risk minimization \sep
  multicalibration \sep
  AI-generated text detection
\end{keywords}

\maketitle

\section{Introduction}

The Voight-Kampff 2026 Task at PAN at CLEF 2026 focused on binary AI-generated (AIG) text detection: identifying if a text came from a large language model (LLM) or a human \cite{bevendorff:2026a, froebe:2023b, bevendorff:2026b, bevendorff:2026c}.
Despite many classifiers achieving high performance when faced with texts that mirror their training distributions, AIG text detectors struggle to generalize to out-of-distribution (OOD) texts \cite{thorat2025dactyl}. 
We suspect that AIG text detection datasets may contain spurious correlations --- e.g., texts that contain unintentional artifacts that coincidentally map to AI-generated content or human-written content. Empirically, for example, we observe that for the RAID benchmark test set, 
a 4 million parameter (BERT-tiny) model achieved a competitive AUROC (0.967) score against larger classifiers\cite{dugan2024raid}. On another AIG text detection dataset, the DACTYL test set, BERT-tiny achieves an AUROC score of 0.99 \cite{thorat2025dactyl}. These results suggest that AIG text detection datasets may contain superficial features that allow for smaller models to obtain misleadingly high scores. A naive way to combat dataset-specific issues is to combine datasets. However, this approach doesn't always help machine-learning models and in some cases \textit{reinforces} a model's dependency on these spurious correlations \cite{compton2023more}.

Previous work to address these issues include hard mining --- retraining classifiers on texts they struggled with --- has yielded some success in improving robustness \cite{emi2024technical}. However, this technique doesn't necessarily distinguish between texts that are genuinely difficult or simply noisy texts (e.g., a corrupted/hallucinated generation).
This lack of distinguishability leads to another issue --- classifier's scores (scaled from 0 to 1, where 0 means human-written and 1 means AIG) aren't necessarily calibrated, meaning the score itself may not be a reliable signal for evaluating a classifier's confidence on an example. Additionally, a single (even calibrated) score doesn't indicate a model's uncertainty on an input.
Finally, a majority of BERT-based classifiers use empirical risk minimization (ERM): the training loop optimizes cross-entropy loss or its variants, such as focal loss. Empirical risk minimization, while successful in many classification tasks and domains, tends to encourage shortcut-learning, leading to classifiers struggling to generalize to OOD texts \cite{korakakis2025mitigating}.

To address these issues, we contribute the following parts of our system:
\begin{itemize}
  \item We construct DACTYLv2, an expansion of the original DACTYL dataset. 
  \item To better select texts for training, we train weak BERT-tiny models with a Bayesian neural network (BNN) classifier head (we refer to this architecture as Bayesian BERT-tiny), which can output different scores for a given text, allowing us to compute $\mu$ (mean score) and $\sigma$ (standard deviation). Using these features, we can filter out texts from different datasets to use for training. These filtering steps allow us to blend different datasets together for one final training set.
  \item Instead of traditional ERM, we use empirical X-risk minimization (EXM), specifically, two-way partial AUROC optimization via the LibAUC library \cite{libauc2023}. EXM has shown promising results in generalizability, especially with larger models \cite{thorat2025dactyl}.
  \item We produce three different classifiers: a ModernBERT-large and a DeBERTa-V3-large classifier \cite{modernbert, he2021deberta}. For our third classifier, we use the weak Bayesian BERT-tiny classifiers' uncertainties ($\sigma$) as features to calibrate ModernBERT-large via MCGrad \cite{tax2026mcgrad}.
  
\end{itemize}

\section{Datasets Used}

\subsection{Training}

\subsubsection{DACTYLv2}

Building upon the work in \citep{thorat2025dactyl}, we construct the second version of DACTYL, which we denote as DACTYLv2 --- we refer to DACTYL-complete as DACTYLv1 and DACTYLv2 combined. We create all AIG texts via one-shot prompting; we gave the LLM a human-written example to base its generation from. 
We included the following domains in DACTYLv2:
\begin{itemize}
  \item blogs \citep{schler2006bac}
  \item web articles \citep{zhang2024finefineweb}
  \item Medium articles \citep{kaggle190kMedium}
  \item movie scripts \citep{saxenakeller2024moviesum}
  \item NBER economic paper abstracts \citep{githubGitHubLedwindranber}
  \item Yelp reviews \citep{kaggleYelpDataset}
\end{itemize} 

We use the following models to construct the dataset. 

\begin{itemize}
  \item Claude Haiku 4.5 \cite{haiku45}
  \item Claude Sonnet 4.6 \cite{sonnet46}
  \item GPT 5.4 \cite{gpt54}
  \item GLM 5.1 \cite{glm5team2026glm5vibecodingagentic}
  \item MiniMax M2.5 \cite{minimax2026minimaxm2seriesminiactivations}
  \item Mercury M2 \cite{labs2025mercuryultrafastlanguagemodels}
\end{itemize}

We used mirror prompting to create LLM prompts as used in \citep{emi2024technical} for the blogs, web articles, and Medium articles. We provided the GPT 5.4 Nano model with the human article and asked it to generate a possible prompt that could have inspired the article \cite{gpt54nano}. For movie scripts, given two randomly selected movie scripts, we take a random excerpt of 4,096 characters from each script. We prompt the LLM to rewrite the first random excerpt by mimicking the style of the second random excerpt. For NBER abstracts, we have the LLM copy the style of a random abstract and generate a new abstract given only the title of another randomly selected paper title. For Yelp reviews, we take a randomly selected business, its industry, and one random review and ask the LLM to generate another review using the selected review as a style reference. 
For Yelp reviews only, we also included texts generated by Kimi-K2.5 \cite{kimiteam2026kimik25visualagentic}. After manual inspection of Kimi-K2.5's generations, we observed that the specific model struggled with style mimicking compared to other LLMs, so we did not use the model for other domains. We report aggregated counts by generator (LLM or human) and domain in Table \ref{tab:model_text_counts} after deduplication. 

For the open-source LLMs, we used the models provided by the TogetherAI and FireworksAI platforms \cite{togetherTogetherNative, fireworksFireworksFastest}. 

\begin{table}[h!]

\centering
\caption{DACTYLv2 Text Counts by Domain/Model across all splits}

\begin{tabular}{lrrrrrrr}
\toprule
\textbf{Model} & \textbf{NBER} & \textbf{Yelp} & \textbf{Web} & \textbf{Blogs} & \textbf{Medium} & \textbf{Scripts} & \textbf{Total} \\
 & \textbf{Abstracts} & & & & \textbf{Articles} & \textbf{(Movie)} & \textbf{Texts} \\
\midrule
GLM 5.1                         & 1{,}400 & 1{,}400 &   160 & 1{,}150 &  5{,}996 & 275 & 10{,}381 \\
GPT 5.4                                 & 1{,}400 & 1{,}400 &   160 & 1{,}150 &  6{,}000 & 275 & 10{,}385 \\
Sonnet 4.6                     & 1{,}399 & 1{,}400 &   158 & 1{,}143 &  5{,}994 & 274 & 10{,}368 \\
Haiku 4.5              & 1{,}400 & 1{,}400 &   160 & 1{,}150 &  6{,}000 & 275 & 10{,}385 \\
Mercury 2                                & 1{,}400 & 1{,}397 &   159 & 1{,}076 &  5{,}985 & 274 & 10{,}291 \\
Kimi-K2.5    &       0 & 1{,}387 &     0 &       0 &        0 &   0 &  1{,}387 \\
Qwen3.5-397B-A17B                   & 1{,}400 & 1{,}337 &   160 & 1{,}150 &  5{,}986 & 275 & 10{,}308 \\
MiniMax-M2.5                   & 1{,}388 & 1{,}396 &   160 & 1{,}133 &  5{,}954 & 229 & 10{,}260 \\
Human                                    & 30{,}600 & 90{,}078 & 21{,}123 & 97{,}727 & 156{,}362 & 2{,}200 & 398{,}090 \\
\midrule
\textbf{Total}                      & \textbf{40{,}387} & \textbf{101{,}198} & \textbf{22{,}240} & \textbf{105{,}679} & \textbf{198{,}277} & \textbf{4{,}077} & \textbf{471{,}858} \\
\bottomrule
\end{tabular}
  \label{tab:model_text_counts}
\end{table}

\subsubsection{External Datasets}
Since DACTYL-complete only includes one-shot generations, we included texts from other existing datasets. Augmenting our training set also ensures that our final classifiers don't overfit to DACTYL-specific artifacts unintentionally.  We selected LLMTrace (English only) \citep{tolstykh2025llmtrace}, Personagen and Stack Exchange \citep{gugliotta2025personagen,huggingfaceHuggingFaceGECLMStackExchange_Mar2023Datasets}, MAGA-Bench (English only) \cite{song2026maga}, and DetectRL \cite{wu2024detectrl} as additional \textit{candidate} training datasets to complement DACTYL-complete. 
LLMTrace's English split includes multiple prompt types, such as updating a human-written text as well as mixed texts. Personagen includes a specific prompt style --- LLMs act as a specific human persona before generating their responses. MAGA-Bench includes generations from different LLM ``alignments'' (e.g., role-playing, black-box prompt optimization, etc.).
DetectRL includes mixed LLM-generated texts where a single AIG text may have contributions from different LLMs. Since Personagen only includes AIG texts, we complement it with a subset of Stack Exchange posts. For each community (that wasn't a meta community), we randomly sampled up to 10,000 posts before November 1, 2022.  

\subsection{Calibration}
We perform an additional layer of calibration for one of our models (ModernBERT-large), given that the scoring metrics for PAN CLEF 2026 include Brier score loss. 
We use the following datasets listed in Table \ref{tab:calibration_counts}. For each dataset, we take a random sample of 2,000 texts, followed by deduplication, to avoid a larger dataset from dominating our calibration set. Due to an imbalance of AIG texts over human texts, we include samples from human-only datasets: CC-News and Stack Exchange. We did not perform calibration on DeBERTa-V3-large since it already obtained strong performance on the calibration set. 
\begin{table}[h!]
\centering
\caption{Calibration counts by dataset and generator.}
\label{tab:calibration_counts}
\begin{tabular}{lrrr}
\toprule
\textbf{Dataset} & \textbf{Total Texts} & \textbf{AI Texts} & \textbf{Human Texts} \\
\midrule
DACTYLv1 (validation)         & 1{,}997 & 738   & 1{,}259 \\
Personagen   & 2{,}000 & 2{,}000 & 0     \\
AIGC Text Bank \cite{wang2026reasoningawareaigcdetectionalignment}         & 2{,}000 & 1{,}913 & 87    \\
AuthorAware                & 2{,}000 & 2{,}000 & 0     \\
OpenTuringBench \cite{openturingbench}           & 1{,}995 & 1{,}995 & 0     \\
DACTYLv2 (validation)      & 2{,}000 & 469   & 1{,}531 \\
PAN-CLEF 2025 (training)\cite{bevendorff2025overview}         & 2{,}000 & 1{,}252 & 748   \\
CC News \cite{ccnews}              & 1{,}961 & 0     & 1{,}961 \\
NY Times AI \cite{nytimesai}              & 2{,}000 & 1{,}713 & 287   \\
Stack Exchange & 2{,}000 & 0     & 2{,}000 \\
SAGE \cite{sage}                     & 1{,}985 & 1{,}523 & 462   \\
LLMTrace (English, validation)  & 2{,}000 & 1{,}210 & 790   \\
Detect AI \cite{detectaicalibration}              & 2{,}000 & 1{,}346 & 654   \\
\midrule
\textbf{Total}            & \textbf{25{,}938} & \textbf{16{,}159} & \textbf{9{,}779} \\
\bottomrule
\end{tabular}
\end{table}

\subsection{Testing}
We evaluate our models on numerous test sets to evaluate generalizability since an updated validation set was not released this year, in contrast to previous years \cite{bevendorff2025overview}. We use the following datasets listed in Table \ref{tab:test_sets}. We included ID test sets (DACTYL-complete, MAGA-Bench, LLMTrace) as well as OOD test sets. 

\begin{table}[h!]
\centering
\caption{External test sets used.}
\label{tab:test_sets}
\begin{tabular}{lll}
\toprule
\textbf{Dataset} & \textbf{Domain Count} & \textbf{LLMs Used} \\
\midrule
BEEMO\cite{artemova2025beemobenchmarkexperteditedmachinegenerated} & 5 & 10 \\
CoCONUTS\cite{chen2025coconutsconcentratingcontentneglecting}     & 6 & 7 \\
DACTYL-complete (DACTYLv2 and DACTYLv1 \cite{thorat2025dactyl})                                     & 12 &   25      \\
DetectRL                                       & 4  & 4      \\
Dolly-Cosmopedia \cite{DatabricksBlog2023DollyV2,benallal2024cosmopedia} & 21 & 1 \\
LLMTrace  (English)                                    & 9  &  19     \\
MAGA-Bench  (English)                                   & 10  &  12    \\
OriginalityAI \cite{githubGitHubOriginalityAIAIdetectorresearchtool} & Not specified    &       4   \\
PAN 2025 Val \cite{bevendorff2025overview}                              & 3    &  22  \\
Personagen + StackExchange                                   & 4     & 6    \\
RealDet (English)\cite{zhu-etal-2025-reliably}          & 15  &    22    \\
UChicago \cite{jabarian2025artificial}         & 6    &   4\\
\bottomrule
\end{tabular}
\end{table}

\section{Data Filtering}

\subsection{Bayesian BERT-Tiny Classifiers}

To get an estimate on dataset generalizability, we train small classifiers on a sample of these datasets. 
Each smaller classifier is a BERT-tiny model with a Bayesian variant of a linear layer as the classification head \cite{berttiny1, berttiny2,blitzbayesian}. 
The Bayesian head's input dimensions are equivalent to the hidden size of the BERT-tiny model, and the output dimension size is 1. Bayesian neural networks (BNNs) allow for the estimation 
of $\sigma$ (standard deviation) of a prediction on a specific input. This uncertainty modeling comes from how a BNN represents its internal weights --- rather than having individual scalars,
the BNN replaces each scalar with a distribution \cite{blundell2015weightuncertaintyneuralnetworks}. During inference time, we sample a set of parameters to use for a prediction. Upon doing $n$ samples, we can compute the ($\mu$, $\sigma$), or the mean score and standard deviation. 

Training a Bayesian BERT-tiny for classification is nearly identical to a standard BERT model's training, except it optimizes the ELBO (evidence lower bound) loss. We can define ELBO loss as

\begin{equation}
\text{ELBO} = L + \frac{\beta}{b}\text{KL}\left[q(w|\theta) || P(w)\right]
\end{equation}

where $L$ is our loss function of choice, $q(w|\theta)$ is our variational approximation to the posterior, and $P(w)$ is our prior. $\theta$ are the parameters that characterize the distribution of weights $w$ for our model.
The KL divergence term prevents the model's weights from moving too far away from our prior. The fraction $\beta/b$ is a scaling factor to avoid the KL term from dominating the ELBO loss. $\beta$ is a positive value (for our experiments, we set it to 1) and $b$ is the mini-batch size during training.

For each of the five training sets, we train a Bayesian BERT-tiny classifier on a random sample of the training set. We use the two-way partial AUROC loss function as $L$ (the DRO-KL variant also used in \cite{thorat2025dactyl}), a batch size of 64, a learning rate of $1 \times 10^{-4}$, and a weight decay of $0.01$. We train each classifier for one epoch. Since the two-way partial AUROC loss needs at least one positive sample per mini-batch, we set the proportion of positive samples equal to 0.5 for each mini-batch. 
Due to each of the five datasets possessing different distributions, we manually tuned the hyperparameters so that the classifier's uncertainty increased on OOD texts. We observed that the random sample size (defined as fraction $f$) and the initial posterior ($\rho$, which can be transformed into the standard deviation of weights via $\sigma = \log{(1 + \exp{\rho})}$) influenced the model's uncertainty significantly. 
For each Bayesian BERT-tiny model, we assign some of the classification head hyperparameters to fixed values, according to the BLiTZ documentation \cite{blitzbayesian}:
\begin{itemize}
  \item prior $\sigma_1$ = 0.1
  \item prior $\sigma_2$ = 1.5
  \item prior $\pi$ = 0.5
  \item initial posterior $\mu$ = 0
\end{itemize}

\begin{table}[h!]
\centering
\caption{Bayesian BERT-tiny training configurations.}
\begin{tabular}{lccc}
\toprule
\textbf{Dataset} & \textbf{Initial Posterior $\rho$} & \textbf{$f$} & \textbf{Training Sample Count}\\
\midrule
DACTYL-complete      & $-3$ & 1/5 & 167,099 \\
DetectRL      & $-3$ & 1 & 205,914 \\
MAGA-Bench          & $-3$ & 1/5 & 71,903 \\
Personagen+SE    & $-3$ & 1/80 & 25,565 \\
LLMTrace & $-2$ & 1  & 173,511\\
\bottomrule
\end{tabular}
\label{tab:training-configs}
\end{table}

We report initial AUROC scores by classifier and test set to evaluate discrimination ability between AI and human texts in Table \ref{tab:bayesian-results}. 

\begin{table}[h!]

\centering
\caption{AUROC scores by Bayesian BERT-tiny classifier.}
\begin{tabular}{lrrrrr}
\toprule
Train Dataset & DACTYL-complete & DetectRL & LLMTrace & MAGA-Bench & Personagen+SE \\
\midrule
DACTYL-complete & 0.9419 & 0.7725 & 0.8692 & 0.9429 & 0.9819 \\
DetectRL & 0.7014 & 0.9839 & 0.7324 & 0.6438 & 0.9702 \\
LLMTrace & 0.8361 & 0.8553 & 0.9633 & 0.9541 & 0.9757 \\
MAGA-Bench & 0.8205 & 0.7881 & 0.8761 & 0.9906 & 0.9978 \\
Personagen+SE & 0.6943 & 0.7597 & 0.8302 & 0.8745 & 0.9999 \\
\bottomrule
\end{tabular}
\label{tab:bayesian-results}
\end{table}

\subsection{Filtering Process}

We observe in Table \ref{tab:bayesian-results} that DACTYL-complete, LLMTrace, and MAGA-Bench are among the best in cross-dataset generalizability. We focus on those datasets and use our classifiers to either ``keep'' or ``reject'' (or ``abstain'' from a decision) texts to use for our final training set. There are four ``quadrants'' of classification outcomes for our Bayesian BERT-tiny models:
\begin{itemize}
    \item correct but uncertain
    \item correct and certain
    \item incorrect but uncertain
    \item incorrect and certain
\end{itemize}

The first three outcomes are acceptable for a model to train on --- a correct classification indicates that a classifier has the \textit{capacity} to learn that sample, regardless of the certainty. An incorrect but uncertain classification for a sample suggests a difficult case --- the classifier is aware that its prediction is possibly wrong.  We discard samples in the fourth and final case, incorrect and certain. The intuition comes from dataset cartography: \citet{datasetcartography} demonstrated that some samples exist in the ``hard-to-learn'' space, which are samples that the model assigns low probability scores to the true class consistently --- these ``hard-to-learn'' samples map roughly to the ``incorrect and certain'' quadrant. While this space can include mislabeled data points, it also includes samples that may be genuine outliers. Given that the two-way partial AUROC loss function focuses on the harder positive (AI) and negative (human) samples, this makes the loss function more sensitive to texts that have high loss and increases the probability of overfitting to outliers \cite{libaucLibauclossesx2014, libauc2023, zhu2022auc}.  

We define the voting process for an individual classifier as follows. For a given text $t$, we compute the standard deviation ($\sigma_t$) and mean score ($\mu_t$, number of runs = 5 to minimize inference time) for the Bayesian BERT-tiny classifier $C_D$, where $D$ is the training dataset. We refer to the distribution of standard deviations on the \textit{test} set for $D$ for classifier $C_D$ as $S_D$. For example, with DACTYL-complete, we compute $S_{\text{DACTYL-complete}}$ by computing the individual $\sigma_t$ values using the Bayesian BERT-tiny classifier trained on DACTYL-complete's subset --- since we want a stable distribution, we do 30 runs for each text in the test set. We can compute the percentile of $\sigma_t$ relative to $S_D$ as $P_t$. A higher $P_t$ indicates a larger value of $\sigma_t$, which implies the model is more uncertain --- conversely, a lower $P_t$ signals high confidence. We convert $\mu_t$ to label $\hat{y}$ (1 or AI-generated if $\mu_t >$  0.5 else 0) to compare to the true label $y$. Using this information, we can then see how  $C_D$ votes on $t$, as described in Table \ref{tab:filtering}. 

\begin{table}[h!]
\centering
\caption{Criteria for voting on text $t$ using classifier $C_D$.}
\begin{tabular}{lc}
\toprule
\textbf{Criteria} & \textbf{Decision} \\
\midrule
(1) $P_t < 5$ or $P_t > 95$ & Abstain \\
(2) $5\leq P_t \leq 95$ and $P_t \leq 50$ and $\hat{y} \neq y$ & Reject $t$ \\
(3) Criteria (1) and (2) are not met & Keep $t$ \\
\bottomrule
\end{tabular}
\label{tab:filtering}
\end{table}

Criterion 1 is a case where the classifier's uncertainty signal appears to be struggling --- extremely high confidence ($\sigma_t$ is less than the 5th percentile) or extreme uncertainty ($\sigma_t$ is greater than the 95th percentile). A classifier abstaining from a text with significantly high uncertainty is reasonable; it hasn't seen this text before to make a reliable vote. However, exceptional overconfidence implies that the classifier's output may be miscalibrated.  

Criterion 2 is the ``incorrect and certain'' sample --- this is a sample where the classifier is relatively confident ($P_t$ is less than the 50th percentile) but makes an incorrect prediction. Criterion 3 is the ``keep'' vote and occurs when the first two criteria aren't met. 

We use the ensemble of classifiers' votes to decide whether or not to keep $t$. $t$ is ultimately kept if at least three of the five classifiers vote to keep and no more than one classifier rejects $t$. For example, if three classifiers vote to keep $t$, but the remaining two classifiers reject it, then we discard $t$. However, if three classifiers vote to keep $t$, and one of the remaining classifiers abstains while the other rejects, we choose to keep $t$ for the final training set. This rejection mechanism helps to minimize the chances of an unusual outlier text making it to the training set.

We report final dataset sizes in Table \ref{tab:filtering_results}. The ``Rejected'' column refers to instances where a text received three ``keep'' votes and two ``reject'' votes.  

\begin{table}[h!]
\centering
\caption{Filtering results per dataset. $\Delta$ denotes the change (percentage points) in the proportion of AIG texts after filtering. The rejected column denotes the number of texts that received three keep votes and two reject votes.}
\begin{tabular}{lrrrrr}
\toprule
\textbf{Train Dataset} & \textbf{Initial $N$} & \textbf{Kept $N$} & \textbf{\% Kept} & \textbf{Rejected} & \textbf{AIG text\% ($\Delta$)} \\
\midrule
DACTYL-complete        & 826,891   & 727,235   & 87.9 & 1,802 & $17.0\% \to 15.4\% (-1.6)$ \\
LLMTrace & 173,511   & 146,908   & 84.7 &   734 & $60.8\% \to 57.5\% (-3.3)$ \\
MAGA-Bench          & 359,518   & 297,994   & 82.9 & 1,387 & $83.4\% \to 81.7\% (-1.7)$ \\
\bottomrule
\end{tabular}
\label{tab:filtering_results}
\end{table}

\section{Model Training}

\subsection{Pre-trained Model Fine-tuning}
We train a (non-Bayesian) DeBERTa-V3-large and ModernBERT-large model on our filtered superset, which consists of all kept texts from the three datasets. We optimize the two-way partial AUROC loss function directly (we don't use ELBO as our objective, as the classification head is not a BNN). 
We use a learning rate of $1\times 10^{-5}$, a batch size of 16, a sampling rate of 0.5, and one epoch for both models.

\subsection{Multicalibration with MCGrad}
For ModernBERT-large, we also perform post-hoc multicalibration with MCGrad using our calibration set \cite{tax2026mcgrad}. Our MCGrad model takes in four inputs --- the transformed score $s$ from the ModernBERT-large model and the individual uncertainties ($\sigma$, 5 runs) for three Bayesian BERT-tiny classifiers (DACTYL, LLMTrace, MAGA-Bench).

We compute $s$ via temperature scaling by transforming the raw probability score $s_r$ into a logit from ModernBERT-large given temperature $T$ in Equation \ref{eqn:scaling}. 

\begin{align}
 \label{eqn:scaling}
  \text{logit}(s_r) &= \log \frac{\operatorname{clip}(s_r,\, 10^{-6},\, 1 - 10^{-6})}{1 - \operatorname{clip}(s_r,\, 10^{-6},\, 1 - 10^{-6})} \\[6pt]
  s &= \frac{1}{1 + \exp\!\left(\frac{-\text{logit}(s_r)}{T}\right)}
\end{align}

We determine $T$ (which we found $T = 2.14555$) by minimizing the log loss between the logits values and the target $y$ values using SciPy's \verb|minimize_scalar| function on the calibration set \cite{2020SciPy-NMeth}. 

We note that MCGrad's algorithm allows for correction, meaning that individual predictions might change significantly.

\section{Evaluation \& Analysis}

For testing on external datasets, we use the same approach described by the workshop organizers \cite{bevendorff:2026a}. We use the following five metrics and the mean of all five metrics (which we denote as the overall score):

\begin{itemize}
  \item AUROC score
  \item micro-$F_1$ score: the harmonic mean between precision and recall
  \item Brier Score: 1 minus the mean squared error between the predicted probability and true label
  \item C@1: modified accuracy that ignores cases where predicted probability = 0.5
  \item $F_{0.5u}$: modified $F_{0.5}$ where abstained cases (predicted probability = 0.5) are considered as false negatives 
\end{itemize}

\begin{table}[h!]
\centering
\caption{ModernBERT-large test set results.}
\begin{tabular}{lrrrrrr}
\toprule
dataset & AUROC & $F_1$ & C@1 & Brier Score & $F_{0.5u}$ & Overall \\
\midrule
BEEMO & 0.8560& 0.9329& 0.8806& 0.9032& 0.9322& 0.9010\\
CoCONUTS & 0.9792& 0.9703& 0.9473& 0.9602& 0.9721& 0.9658\\
DACTYL-complete & 0.9903& 0.9603& 0.9747& 0.9793& 0.9761& 0.9761\\
DetectRL & 0.9370& 0.8910& 0.8852& 0.9054& 0.8992& 0.9036\\
Dolly-Cosmopedia & 0.9958& 0.9161& 0.8953& 0.9202& 0.8726& 0.9200\\
LLMTrace & 0.9909& 0.9730& 0.9675& 0.9735& 0.9797& 0.9769\\
MAGA-Bench & 0.9992& 0.9967& 0.9945& 0.9955& 0.9964& 0.9965\\
OriginalityAI & 0.9322& 0.7465& 0.7820& 0.8371& 0.8273& 0.8250\\
PAN 2025-Validation & 0.9984& 0.9159& 0.8819& 0.9174& 0.8727& 0.9172\\
Personagen+SE & 1.0000& 0.9825& 0.9799& 0.9822& 0.9723& 0.9834\\
RealDet & 0.9540& 0.9158& 0.9178& 0.9314& 0.9284& 0.9295\\
UChicago & 0.9784& 0.9365& 0.9181& 0.9365& 0.9653& 0.9470\\
\midrule
PAN 2025 Test & 0.984 & 0.925 & 0.89 & 0.921 & 0.899 & 0.924 \\
PAN 2026 Test & 0.994 & 0.962 & 0.941 & 0.958 & 0.946 & 0.96 \\
PAN 2026 ELOQUENT Test & 0.927 & 0.929 & 0.873 & 0.91 & 0.961 & 0.92\\

\bottomrule
\end{tabular}
\label{tab:modernbert-large-results}
\end{table}

\begin{table}[h!]
\centering
\caption{ModernBERT-large with MCGrad post-hoc calibration test set results.}
\begin{tabular}{lrrrrrr}
\toprule
dataset & AUROC & $F_1$ & C@1 & Brier Score & $F_{0.5u}$ & Overall \\
\midrule
BEEMO & 0.8619& 0.9337& 0.8826& 0.9112& 0.9350& 0.9049\\
CoCONUTS & 0.9797& 0.9667& 0.9420& 0.9593& 0.9785& 0.9652\\
DACTYL-complete & 0.9893& 0.9509& 0.9688& 0.9757& 0.9685& 0.9707\\
DetectRL & 0.9384& 0.8869& 0.8830& 0.9085& 0.9054& 0.9044\\
Dolly-Cosmopedia & 0.9931& 0.9053& 0.8804& 0.9110& 0.8573& 0.9094\\
LLMTrace & 0.9906& 0.9723& 0.9666& 0.9722& 0.9771& 0.9758\\
MAGA-Bench & 0.9988& 0.9955& 0.9925& 0.9938& 0.9949& 0.9951\\
OriginalityAI & 0.9303& 0.7422& 0.7860& 0.8424& 0.8462& 0.8294\\
PAN 2025-Validation & 0.9986& 0.9559& 0.9407& 0.9551& 0.9325& 0.9566\\
Personagen+SE & 1.0000& 0.9909& 0.9897& 0.9887& 0.9856& 0.9910\\
RealDet & 0.9586& 0.9120& 0.9152& 0.9353& 0.9327& 0.9308\\
UChicago & 0.9781& 0.9332& 0.9140& 0.9394& 0.9629& 0.9455\\
\midrule 
PAN 2025 Test & 0.984 & 0.95 &  0.928 & 0.946 & 0.939 &  0.949 \\
PAN 2026 Test & 0.993 & 0.975 & 0.962 & 0.97 & 0.969 & 0.974 \\
PAN 2026 ELOQUENT Test &  0.943 &  0.911 & 0.844 & 0.899 &   0.959 & 0.911\\
\bottomrule
\end{tabular}
\label{tab:modernbert-large-results-calibrated}
\end{table}

\begin{table}[h!]
  \centering
\caption{DeBERTa-V3-large test set results.}
\begin{tabular}{lrrrrrr}
\toprule
dataset & AUROC & $F_1$ & C@1 & Brier Score & $F_{0.5u}$ & Overall \\
\midrule
BEEMO & 0.8780 & 0.9302 & 0.8785 & 0.9041 & 0.9418 & 0.9065 \\
CoCONUTS & 0.9819 & 0.9518 & 0.9180 & 0.9296 & 0.9785 & 0.9519 \\
DACTYL-complete & 0.9846 & 0.9524 & 0.9698 & 0.9747 & 0.9702 & 0.9703 \\
DetectRL & 0.9465 & 0.8759 & 0.8756 & 0.8940 & 0.9124 & 0.9009 \\
Dolly-Cosmopedia & 0.9952 & 0.9274 & 0.9107 & 0.9386 & 0.8905 & 0.9325 \\
LLMTrace & 0.9903 & 0.9738 & 0.9686 & 0.9739 & 0.9842 & 0.9782 \\
MAGA-Bench & 0.9992 & 0.9966 & 0.9944 & 0.9956 & 0.9974 & 0.9967 \\
OriginalityAI & 0.9213 & 0.6551 & 0.7425 & 0.7741 & 0.8227 & 0.7831 \\
PAN 2025-Validation & 0.9985 & 0.9917 & 0.9894 & 0.9919 & 0.9951 & 0.9933 \\
Personagen+SE & 1.0000 & 0.9973 & 0.9970 & 0.9976 & 0.9958 & 0.9975 \\
RealDet & 0.9810 & 0.9389 & 0.9418 & 0.9518 & 0.9670 & 0.9561 \\
UChicago & 0.9817 & 0.8989 & 0.8756 & 0.8981 & 0.9563 & 0.9221 \\
\midrule
PAN 2025 Test & 0.987 & 0.975 & 0.965 & 0.969 & 0.987 & 0.977 \\
PAN 2026 Test & 0.989 & 0.843 & 0.797 & 0.852 & 0.93 & 0.882 \\
PAN 2026 ELOQUENT Test &  0.982 & 0.865 & 0.774 &0.822 & 0.941 & 0.877\\

\bottomrule
\end{tabular}

\end{table}

\begin{table}[h!]
\centering
\caption{Final leaderboard results for the PAN 2026 test set for teams that achieved higher overall scores compared to the zero-knowledge baseline.}
\begin{tabular}{llrrrrrrrr}
\toprule
Rank & Team & AUROC & $F_1$ & C@1 & Brier Score & $F_{0.5u}$ & Overall \\
\midrule
1 & suspiciously-coherent & 0.989 & 0.976 & 0.965 & 0.965 & 0.980 & 0.975 \\
2 & \textbf{DACTYL (ModernBERT-large + MCGrad)} & 0.993 & 0.975 & 0.962 & 0.970 & 0.969 & 0.974 \\

3 & plagiarism-detector-com-reg2 & 0.997 & 0.955 & 0.949 & 0.958 & 0.982 & 0.968 \\
\midrule
- & \textit{Baseline mdok (PAN 2025 Winner)} & 0.995 & 0.963 & 0.946 & 0.951 & 0.984 & 0.968 \\
\midrule
4 & dsgt-pan-26 & 0.981 & 0.907 & 0.873 & 0.885 & 0.961 & 0.921 \\
5 & ydong & 0.930 & 0.928 & 0.886 & 0.915 & 0.896 & 0.911 \\
6 & turingteam & 0.957 & 0.851 & 0.846 & 0.882 & 0.925 & 0.892 \\
7 & writerslogic-inc & 0.891 & 0.902 & 0.853 & 0.891 & 0.899 & 0.887 \\
8 & a-a-detection & 0.852 & 0.885 & 0.826 & 0.848 & 0.880 & 0.858 \\
\midrule
- & \textit{Baseline PPMd} & 0.783 & 0.865 & 0.783 & 0.810 & 0.828 & 0.814 \\
\midrule
9 & ak-ys & 0.911 & 0.774 & 0.723 & 0.754 & 0.891 & 0.811 \\
10 & ikr3 & 0.814 & 0.856 & 0.749 & 0.779 & 0.788 & 0.797 \\
11 & radar & 0.772 & 0.800 & 0.722 & 0.805 & 0.839 & 0.788 \\
12 & sinai & 0.624 & 0.856 & 0.748 & 0.771 & 0.788 & 0.757 \\
13 & egrantha-ai & 0.500 & 0.856 & 0.749 & 0.749 & 0.788 & 0.728 \\
\midrule
- & \textit{Baseline Zero-knowledge} & 0.500 & 0.856 & 0.749 & 0.749 & 0.788 & 0.728 \\
\bottomrule
\end{tabular}
\label{tab:pan2026-lb}
\end{table}

Both uncalibrated models (ModernBERT-large and DeBERTa-V3-large) have relatively high AUROC scores, even across OOD test sets --- for example, we excluded DetectRL during training, but both models achieved an AUROC score of at least 0.93, highlighting some discrimination capacity. ModernBERT-large has an advantage in the overall score; on the PAN test sets, it achieves an overall score of at least 0.92. However, DeBERTa-V3-large shows more robustness in terms of AUROC score --- on the ELOQUENT test set, it achieves an AUROC score of 0.982 in contrast with ModernBERT-large's 0.927. Our results indicate that a high discrimination capacity can hide miscalibrated scores --- a model that achieves high AUROC is not guaranteed to have outputs that are calibrated. This discrepancy highlights the primary advantage of evaluating across multiple metrics. 

ModernBERT-large with MCGrad yields improvements in the overall score in the PAN 2025 and 2026 test sets compared to the base model. However, it fails to give an improvement on the ELOQUENT test set, suggesting that the multicalibration approach is sensitive to distribution shifts. We confirm this on the 12 external test sets: we see that using the MCGrad and Bayesian BERT-tiny models shows an increase in the overall score for six of the test sets. The increase in score for the PAN test sets (excluding ELOQUENT) might indicate that those test sets represent a distribution shift that our MCGrad model covers.

In comparison to other classifiers, our ModernBERT-large and MCGrad classifier ranks second in overall score and is one of two models to beat the PAN 2025 winning system. The MCGrad system ranks second in AUROC, $F_1$,  C@1, $F_{0.5u}$, and first in Brier score among competing teams.

\section{Conclusion}

 Our work applies Bayesian neural networks (as a classification head) to help select candidate texts to use for downstream training. Additionally, these same classifiers can help improve calibration of larger, non-Bayesian models via their uncertainties (standard deviations). We also point out that both ModernBERT-large and DeBERTa-V3-large have strong generalization capacities given their high AUROC scores across OOD test sets. Using the Bayesian BERT-tiny models' uncertainties as inputs for an MCGrad model demonstrates mixed results on external test sets but an increase in the overall score on the PAN test sets. Note that our Bayesian BERT-tiny hyperparameters for the BNN layer were manually tuned --- we suspect that hyperparameter optimization via some proxy uncertainty metric could lead to better uncertainty estimates, which could improve the MCGrad model. 
 
 Future work might explore applying this Bayesian data filtering technique and EXM on other text classification tasks, such as prompt safety detection.

\section*{Resources Used}

Training Bayesian BERT-tiny models and inference for the classifiers was done on a NVIDIA GB10. We used Lambda's GH200 instances for training ModernBERT-large and DeBERTa-V3-large \cite{lambdaSuperintelligenceCloud}.



\section*{Declaration on Generative AI}
  The author has used Grammarly to perform grammar checking in preparing this paper.


\bibliography{sources.bib}

\begin{thebibliography}{57}
\expandafter\ifx\csname natexlab\endcsname\relax\def\natexlab#1{#1}\fi
\providecommand{\url}[1]{\texttt{#1}}
\providecommand{\href}[2]{#2}
\providecommand{\path}[1]{#1}
\providecommand{\DOIprefix}{doi:}
\providecommand{\ArXivprefix}{arXiv:}
\providecommand{\URLprefix}{URL: }
\providecommand{\Pubmedprefix}{pmid:}
\providecommand{\doi}[1]{\href{http://dx.doi.org/#1}{\path{#1}}}
\providecommand{\Pubmed}[1]{\href{pmid:#1}{\path{#1}}}
\providecommand{\bibinfo}[2]{#2}
\ifx\xfnm\relax \def\xfnm[#1]{\unskip,\space#1}\fi
\bibitem[{Bevendorff et~al.(2026)Bevendorff, Fr{\"o}be, Greiner-Petter, Jakoby, Mayerl, Nakov, Plutz, Ta, Wang, and Zangerle}]{bevendorff:2026a}
\bibinfo{author}{J.~Bevendorff}, \bibinfo{author}{M.~Fr{\"o}be}, \bibinfo{author}{A.~Greiner-Petter}, \bibinfo{author}{A.~Jakoby}, \bibinfo{author}{M.~Mayerl}, \bibinfo{author}{P.~Nakov}, \bibinfo{author}{H.~Plutz}, \bibinfo{author}{M.~Ta}, \bibinfo{author}{Y.~Wang}, \bibinfo{author}{E.~Zangerle},
\newblock \bibinfo{title}{{Overview of PAN 2026: Voight-Kampff Generative AI Detection, Text Watermarking, Multi-author Writing Style Analysis, Generative Plagiarism Detection, and Reasoning Trajectory Detection -- Extended Abstract}},
\newblock in: \bibinfo{editor}{R.~Campos}, \bibinfo{editor}{A.~Jatowt}, \bibinfo{editor}{Y.~Lan}, \bibinfo{editor}{M.~Aliannejadi}, \bibinfo{editor}{C.~Bauer}, \bibinfo{editor}{S.~MacAvaney}, \bibinfo{editor}{Z.~Ren}, \bibinfo{editor}{S.~Verberne}, \bibinfo{editor}{N.~Bai}, \bibinfo{editor}{M.~Mansoury} (Eds.), \bibinfo{booktitle}{Advances in Information Retrieval. 48th European Conference on IR Research (ECIR 2026)}, volume \bibinfo{volume}{16485} of \textit{\bibinfo{series}{Lecture Notes in Computer Science}}, \bibinfo{publisher}{Springer Nature}, \bibinfo{address}{Cham, Switzerland}, \bibinfo{year}{2026}, pp. \bibinfo{pages}{225--232}. \DOIprefix\doi{10.1007/978-3-032-21321-1_32}.
\bibitem[{Fr{\"o}be et~al.(2023)Fr{\"o}be, Wiegmann, Kolyada, Grahm, Elstner, Loebe, Hagen, Stein, and Potthast}]{froebe:2023b}
\bibinfo{author}{M.~Fr{\"o}be}, \bibinfo{author}{M.~Wiegmann}, \bibinfo{author}{N.~Kolyada}, \bibinfo{author}{B.~Grahm}, \bibinfo{author}{T.~Elstner}, \bibinfo{author}{F.~Loebe}, \bibinfo{author}{M.~Hagen}, \bibinfo{author}{B.~Stein}, \bibinfo{author}{M.~Potthast},
\newblock \bibinfo{title}{{Continuous Integration for Reproducible Shared Tasks with TIRA.io}},
\newblock in: \bibinfo{editor}{J.~Kamps}, \bibinfo{editor}{L.~Goeuriot}, \bibinfo{editor}{F.~Crestani}, \bibinfo{editor}{M.~Maistro}, \bibinfo{editor}{H.~Joho}, \bibinfo{editor}{B.~Davis}, \bibinfo{editor}{C.~Gurrin}, \bibinfo{editor}{U.~Kruschwitz}, \bibinfo{editor}{A.~Caputo} (Eds.), \bibinfo{booktitle}{Advances in Information Retrieval. 45th European Conference on {IR} Research ({ECIR} 2023)}, Lecture Notes in Computer Science, \bibinfo{publisher}{Springer}, \bibinfo{address}{Berlin Heidelberg New York}, \bibinfo{year}{2023}, pp. \bibinfo{pages}{236--241}. \URLprefix \url{https://link.springer.com/chapter/10.1007/978-3-031-28241-6_20}. \DOIprefix\doi{10.1007/978-3-031-28241-6_20}.
\bibitem[{Bevendorff et~al.(2026{\natexlab{a}})Bevendorff, Fr{\"o}be, Greiner-Petter, Jakoby, Mayerl, Nakov, Plutz, Potthast, Stein, Ta, Wang, and Zangerle}]{bevendorff:2026b}
\bibinfo{author}{J.~Bevendorff}, \bibinfo{author}{M.~Fr{\"o}be}, \bibinfo{author}{A.~Greiner-Petter}, \bibinfo{author}{A.~Jakoby}, \bibinfo{author}{M.~Mayerl}, \bibinfo{author}{P.~Nakov}, \bibinfo{author}{H.~Plutz}, \bibinfo{author}{M.~Potthast}, \bibinfo{author}{B.~Stein}, \bibinfo{author}{M.~N. Ta}, \bibinfo{author}{Y.~Wang}, \bibinfo{author}{E.~Zangerle},
\newblock in: \bibinfo{editor}{M.~Hagen}, \bibinfo{editor}{M.~Potthast}, \bibinfo{editor}{B.~Stein}, \bibinfo{editor}{P.~Schaer}, \bibinfo{editor}{E.~Zangerle}, \bibinfo{editor}{A.~Barr{\'o}n-Cede{\~n}o}, \bibinfo{editor}{A.~G.~S. de~Herrera}, \bibinfo{editor}{E.~S. Salido}, \bibinfo{editor}{S.~MacAvaney}, \bibinfo{editor}{J.~M. Stru{\ss}} (Eds.), \bibinfo{booktitle}{Experimental IR Meets Multilinguality, Multimodality, and Interaction. Proceedings of the Seventeenth International Conference of the CLEF Association (CLEF 2026)}, Lecture Notes in Computer Science, \bibinfo{publisher}{Springer}, \bibinfo{address}{Berlin Heidelberg New York}, \bibinfo{year}{2026}{\natexlab{a}}.
\bibitem[{Bevendorff et~al.(2026{\natexlab{b}})Bevendorff, Gunti, Karlgren, Fr{\"o}be, Potthast, and Stein}]{bevendorff:2026c}
\bibinfo{author}{J.~Bevendorff}, \bibinfo{author}{R.~R. Gunti}, \bibinfo{author}{J.~Karlgren}, \bibinfo{author}{M.~Fr{\"o}be}, \bibinfo{author}{M.~Potthast}, \bibinfo{author}{B.~Stein},
\newblock \bibinfo{title}{{Overview of the third ``Voight-Kampff'' Generative AI / LLM Detection Task at PAN and ELOQUENT 2026}},
\newblock in: \bibinfo{editor}{A.~Barr{\'o}n-Cede{\~n}o}, \bibinfo{editor}{A.~G.~S. de~Herrera}, \bibinfo{editor}{E.~S. Salido}, \bibinfo{editor}{S.~MacAvaney}, \bibinfo{editor}{J.~M. Stru{\ss}} (Eds.), \bibinfo{booktitle}{Working Notes of CLEF 2026 -- Conference and Labs of the Evaluation Forum}, CEUR Workshop Proceedings, \bibinfo{publisher}{CEUR-WS.org}, \bibinfo{year}{2026}{\natexlab{b}}.
\bibitem[{Thorat and Caines(2025)}]{thorat2025dactyl}
\bibinfo{author}{S.~Thorat}, \bibinfo{author}{A.~Caines},
\newblock \bibinfo{title}{Dactyl: Diverse adversarial corpus of texts yielded from large language models},
\newblock \bibinfo{journal}{arXiv preprint arXiv:2508.00619}  (\bibinfo{year}{2025}).
\bibitem[{Dugan et~al.(2024)Dugan, Hwang, Trhl{\'\i}k, Zhu, Ludan, Xu, Ippolito, and Callison-Burch}]{dugan2024raid}
\bibinfo{author}{L.~Dugan}, \bibinfo{author}{A.~Hwang}, \bibinfo{author}{F.~Trhl{\'\i}k}, \bibinfo{author}{A.~Zhu}, \bibinfo{author}{J.~M. Ludan}, \bibinfo{author}{H.~Xu}, \bibinfo{author}{D.~Ippolito}, \bibinfo{author}{C.~Callison-Burch},
\newblock \bibinfo{title}{Raid: A shared benchmark for robust evaluation of machine-generated text detectors},
\newblock in: \bibinfo{booktitle}{Proceedings of the 62nd Annual Meeting of the Association for Computational Linguistics (Volume 1: Long Papers)}, \bibinfo{year}{2024}, pp. \bibinfo{pages}{12463--12492}.
\bibitem[{Compton et~al.(2023)Compton, Zhang, Puli, and Ranganath}]{compton2023more}
\bibinfo{author}{R.~Compton}, \bibinfo{author}{L.~Zhang}, \bibinfo{author}{A.~Puli}, \bibinfo{author}{R.~Ranganath},
\newblock \bibinfo{title}{When more is less: Incorporating additional datasets can hurt performance by introducing spurious correlations},
\newblock in: \bibinfo{booktitle}{Machine learning for healthcare conference}, \bibinfo{organization}{PMLR}, \bibinfo{year}{2023}, pp. \bibinfo{pages}{110--127}.
\bibitem[{Emi and Spero(2024)}]{emi2024technical}
\bibinfo{author}{B.~Emi}, \bibinfo{author}{M.~Spero},
\newblock \bibinfo{title}{Technical report on the pangram ai-generated text classifier},
\newblock \bibinfo{journal}{arXiv preprint arXiv:2402.14873}  (\bibinfo{year}{2024}).
\bibitem[{Korakakis et~al.(2025)Korakakis, Vlachos, and Weller}]{korakakis2025mitigating}
\bibinfo{author}{M.~Korakakis}, \bibinfo{author}{A.~Vlachos}, \bibinfo{author}{A.~Weller},
\newblock \bibinfo{title}{Mitigating shortcut learning with interpolated learning},
\newblock in: \bibinfo{booktitle}{Proceedings of the 63rd Annual Meeting of the Association for Computational Linguistics (Volume 1: Long Papers)}, \bibinfo{year}{2025}, pp. \bibinfo{pages}{9191--9206}.
\bibitem[{Yuan et~al.(2023)Yuan, Zhu, Qiu, Li, Wang, and Yang}]{libauc2023}
\bibinfo{author}{Z.~Yuan}, \bibinfo{author}{D.~Zhu}, \bibinfo{author}{Z.-H. Qiu}, \bibinfo{author}{G.~Li}, \bibinfo{author}{X.~Wang}, \bibinfo{author}{T.~Yang},
\newblock \bibinfo{title}{Libauc: A deep learning library for x-risk optimization},
\newblock in: \bibinfo{booktitle}{Proceedings of the 29th ACM SIGKDD Conference on Knowledge Discovery and Data Mining}, KDD ’23, \bibinfo{publisher}{ACM}, \bibinfo{year}{2023}, p. \bibinfo{pages}{5487–5499}. \URLprefix \url{http://dx.doi.org/10.1145/3580305.3599861}. \DOIprefix\doi{10.1145/3580305.3599861}.
\bibitem[{Warner et~al.(2025)Warner, Chaffin, Clavi{\'e}, Weller, Hallstr{\"o}m, Taghadouini, Gallagher, Biswas, Ladhak, Aarsen, Adams, Howard, and Poli}]{modernbert}
\bibinfo{author}{B.~Warner}, \bibinfo{author}{A.~Chaffin}, \bibinfo{author}{B.~Clavi{\'e}}, \bibinfo{author}{O.~Weller}, \bibinfo{author}{O.~Hallstr{\"o}m}, \bibinfo{author}{S.~Taghadouini}, \bibinfo{author}{A.~Gallagher}, \bibinfo{author}{R.~Biswas}, \bibinfo{author}{F.~Ladhak}, \bibinfo{author}{T.~Aarsen}, \bibinfo{author}{G.~T. Adams}, \bibinfo{author}{J.~Howard}, \bibinfo{author}{I.~Poli},
\newblock \bibinfo{title}{Smarter, better, faster, longer: A modern bidirectional encoder for fast, memory efficient, and long context finetuning and inference},
\newblock in: \bibinfo{editor}{W.~Che}, \bibinfo{editor}{J.~Nabende}, \bibinfo{editor}{E.~Shutova}, \bibinfo{editor}{M.~T. Pilehvar} (Eds.), \bibinfo{booktitle}{Proceedings of the 63rd Annual Meeting of the Association for Computational Linguistics (Volume 1: Long Papers)}, \bibinfo{publisher}{Association for Computational Linguistics}, \bibinfo{address}{Vienna, Austria}, \bibinfo{year}{2025}, pp. \bibinfo{pages}{2526--2547}. \URLprefix \url{https://aclanthology.org/2025.acl-long.127/}. \DOIprefix\doi{10.18653/v1/2025.acl-long.127}.
\bibitem[{He et~al.(2021)He, Liu, Gao, and Chen}]{he2021deberta}
\bibinfo{author}{P.~He}, \bibinfo{author}{X.~Liu}, \bibinfo{author}{J.~Gao}, \bibinfo{author}{W.~Chen},
\newblock \bibinfo{title}{Deberta: Decoding-enhanced bert with disentangled attention},
\newblock in: \bibinfo{booktitle}{International Conference on Learning Representations}, \bibinfo{year}{2021}. \URLprefix \url{https://openreview.net/forum?id=XPZIaotutsD}.
\bibitem[{Tax et~al.(2026)Tax, Perini, Linder, Haimovich, Karamshuk, Okati, Vojnovic, and Apostolopoulos}]{tax2026mcgrad}
\bibinfo{author}{N.~Tax}, \bibinfo{author}{L.~Perini}, \bibinfo{author}{F.~Linder}, \bibinfo{author}{D.~Haimovich}, \bibinfo{author}{D.~Karamshuk}, \bibinfo{author}{N.~Okati}, \bibinfo{author}{M.~Vojnovic}, \bibinfo{author}{P.~A. Apostolopoulos},
\newblock \bibinfo{title}{{MCGrad: Multicalibration at Web Scale}},
\newblock in: \bibinfo{booktitle}{Proceedings of the 32nd ACM SIGKDD Conference on Knowledge Discovery and Data Mining V.1 (KDD 2026)}, \bibinfo{year}{2026}. \DOIprefix\doi{10.1145/3770854.3783954}.
\bibitem[{Schler et~al.(2006)Schler, Koppel, Argamon, and Pennebaker}]{schler2006bac}
\bibinfo{author}{J.~Schler}, \bibinfo{author}{M.~Koppel}, \bibinfo{author}{S.~Argamon}, \bibinfo{author}{J.~W. Pennebaker},
\newblock \bibinfo{title}{Effects of age and gender on blogging.},
\newblock in: \bibinfo{booktitle}{AAAI spring symposium: Computational approaches to analyzing weblogs}, volume~\bibinfo{volume}{6}, \bibinfo{organization}{Stanford, CA, USA}, \bibinfo{year}{2006}, pp. \bibinfo{pages}{199--205}.
\bibitem[{Zhang et~al.(2024)Zhang, Du, Yu, Wang, Wang, Guo, Zheng, Zhu, Liu, Yue, Liu, Peng, Yao, Yang, Li, Zhang, Liu, Liu, Gao, Chen, Zhou, Liu, Wang, and Huang}]{zhang2024finefineweb}
\bibinfo{author}{G.~Zhang}, \bibinfo{author}{X.~Du}, \bibinfo{author}{Z.~Yu}, \bibinfo{author}{Z.~Wang}, \bibinfo{author}{Z.~Wang}, \bibinfo{author}{S.~Guo}, \bibinfo{author}{T.~Zheng}, \bibinfo{author}{K.~Zhu}, \bibinfo{author}{J.~Liu}, \bibinfo{author}{S.~Yue}, \bibinfo{author}{B.~Liu}, \bibinfo{author}{Z.~Peng}, \bibinfo{author}{Y.~Yao}, \bibinfo{author}{J.~Yang}, \bibinfo{author}{Z.~Li}, \bibinfo{author}{B.~Zhang}, \bibinfo{author}{M.~Liu}, \bibinfo{author}{T.~Liu}, \bibinfo{author}{Y.~Gao}, \bibinfo{author}{W.~Chen}, \bibinfo{author}{X.~Zhou}, \bibinfo{author}{Q.~Liu}, \bibinfo{author}{T.~Wang}, \bibinfo{author}{W.~Huang}, \bibinfo{title}{Finefineweb: A comprehensive study on fine-grained domain web corpus}, \bibinfo{year}{2024}. \URLprefix \url{https://huggingface.co/datasets/m-a-p/FineFineWeb}, \bibinfo{note}{version v0.1.0}.
\bibitem[{Chiusano(2022)}]{kaggle190kMedium}
\bibinfo{author}{F.~Chiusano}, \bibinfo{title}{190k+ {M}edium {A}rticles --- kaggle.com}, \bibinfo{howpublished}{\url{https://www.kaggle.com/datasets/fabiochiusano/medium-articles/data}}, \bibinfo{year}{2022}. \bibinfo{note}{[Accessed 17-05-2026]}.
\bibitem[{Saxena and Keller(2024)}]{saxenakeller2024moviesum}
\bibinfo{author}{R.~Saxena}, \bibinfo{author}{F.~Keller},
\newblock \bibinfo{title}{{M}ovie{S}um: An abstractive summarization dataset for movie screenplays},
\newblock in: \bibinfo{editor}{L.-W. Ku}, \bibinfo{editor}{A.~Martins}, \bibinfo{editor}{V.~Srikumar} (Eds.), \bibinfo{booktitle}{Findings of the Association for Computational Linguistics: ACL 2024}, \bibinfo{publisher}{Association for Computational Linguistics}, \bibinfo{address}{Bangkok, Thailand}, \bibinfo{year}{2024}, pp. \bibinfo{pages}{4043--4050}. \URLprefix \url{https://aclanthology.org/2024.findings-acl.239/}. \DOIprefix\doi{10.18653/v1/2024.findings-acl.239}.
\bibitem[{Edwindra(2025)}]{githubGitHubLedwindranber}
\bibinfo{author}{L.~Edwindra}, \bibinfo{title}{{G}it{H}ub - ledwindra/nber: {S}crape and analyzie {N}{B}{E}{R} working papers. --- github.com}, \bibinfo{howpublished}{\url{https://github.com/ledwindra/nber}}, \bibinfo{year}{2025}. \bibinfo{note}{[Accessed 17-05-2026]}.
\bibitem[{Yelp(2022)}]{kaggleYelpDataset}
\bibinfo{author}{Yelp}, \bibinfo{title}{{Y}elp {D}ataset --- kaggle.com}, \bibinfo{howpublished}{\url{https://www.kaggle.com/datasets/yelp-dataset/yelp-dataset}}, \bibinfo{year}{2022}. \bibinfo{note}{[Accessed 17-05-2026]}.
\bibitem[{Anthropic(2025)}]{haiku45}
\bibinfo{author}{Anthropic}, \bibinfo{title}{{I}ntroducing {C}laude {H}aiku 4.5 --- anthropic.com}, \bibinfo{howpublished}{\url{https://www.anthropic.com/news/claude-haiku-4-5}}, \bibinfo{year}{2025}. \bibinfo{note}{[Accessed 28-05-2026]}.
\bibitem[{Anthropic(2026)}]{sonnet46}
\bibinfo{author}{Anthropic}, \bibinfo{title}{{I}ntroducing {S}onnet 4.6 --- anthropic.com}, \bibinfo{howpublished}{\url{https://www.anthropic.com/news/claude-sonnet-4-6}}, \bibinfo{year}{2026}. \bibinfo{note}{[Accessed 28-05-2026]}.
\bibitem[{OpenAI(2026)}]{gpt54}
\bibinfo{author}{OpenAI}, \bibinfo{title}{{I}ntroducing {G}{P}{T}‑5.4}, \bibinfo{howpublished}{\url{https://openai.com/index/introducing-gpt-5-4/}}, \bibinfo{year}{2026}. \bibinfo{note}{[Accessed 28-05-2026]}.
\bibitem[{GLM-5-Team et~al.(2026)GLM-5-Team, :, Zeng, Lv, Hou, Du, Zheng, Chen, Yin, Ge, Huang, Xie, Zhu, Yin, Wang, Pan, Zeng, Zhang, Wang, Chen, Zhang, Jiao, Guo, Wang, Du, Wu, Wang, Li, Fan, Zhong, Liu, Zhao, Du, Dong, Lu, Shuang-Li, Cao, Liu, Jiang, Chen, Zhang, Huang, Dong, Xu, Wei, An, Niu, Zhu, Wen, Cen, Bai, Qiao, Wang, Wang, Zhu, Liu, Li, Wang, Wen, Huang, Cai, Yu, Li, Hu, Zhang, Zhang, Lin, Yang, Wang, Ai, Zhu, Yi, Chen, Wen, Sun, Zhao, Hu, Zhang, Liu, Zhang, Peng, Tai, Zhang, Liu, Wang, Yan, Ge, Liu, Chu, Zhao, Wang, Zhao, Ren, Wang, Zhang, Gui, Zhao, Li, An, Li, Yuan, Du, Liu, Zhi, Duan, Zhou, Wei, Wang, Luo, Zhang, Sha, Xu, Wu, Ding, Chen, Li, Lin, Ta, Zou, Song, Yang, Tu, Yang, Wu, Zhang, Li, Li, Fan, Qin, Tian, Zhang, Yu, Liang, Kuang, Cheng, Li, Yan, Hu, Ling, Fan, Xia, Zhang, Zhang, Pan, Zou, Zhang, Liu, Wu, Li, Wang, Zhu, Tan, Zhou, Pan, Zhang, Su, Geng, Yan, Tan, Bi, Shen, Yang, Li, Liu, Wang, Li, Wu, Zhang, Duan, Zhang, Liu, Jiang, Yan, Zhang, Wei, Chen, Feng, Yao, Chai, Wang, Zhang, Xu,
  Huang, Wang, Li, Dong, and Tang}]{glm5team2026glm5vibecodingagentic}
\bibinfo{author}{GLM-5-Team}, \bibinfo{author}{:}, \bibinfo{author}{A.~Zeng}, \bibinfo{author}{X.~Lv}, \bibinfo{author}{Z.~Hou}, \bibinfo{author}{Z.~Du}, \bibinfo{author}{Q.~Zheng}, \bibinfo{author}{B.~Chen}, \bibinfo{author}{D.~Yin}, \bibinfo{author}{C.~Ge}, \bibinfo{author}{C.~Huang}, \bibinfo{author}{C.~Xie}, \bibinfo{author}{C.~Zhu}, \bibinfo{author}{C.~Yin}, \bibinfo{author}{C.~Wang}, \bibinfo{author}{G.~Pan}, \bibinfo{author}{H.~Zeng}, \bibinfo{author}{H.~Zhang}, \bibinfo{author}{H.~Wang}, \bibinfo{author}{H.~Chen}, \bibinfo{author}{J.~Zhang}, \bibinfo{author}{J.~Jiao}, \bibinfo{author}{J.~Guo}, \bibinfo{author}{J.~Wang}, \bibinfo{author}{J.~Du}, \bibinfo{author}{J.~Wu}, \bibinfo{author}{K.~Wang}, \bibinfo{author}{L.~Li}, \bibinfo{author}{L.~Fan}, \bibinfo{author}{L.~Zhong}, \bibinfo{author}{M.~Liu}, \bibinfo{author}{M.~Zhao}, \bibinfo{author}{P.~Du}, \bibinfo{author}{Q.~Dong}, \bibinfo{author}{R.~Lu}, \bibinfo{author}{Shuang-Li}, \bibinfo{author}{S.~Cao}, \bibinfo{author}{S.~Liu},
  \bibinfo{author}{T.~Jiang}, \bibinfo{author}{X.~Chen}, \bibinfo{author}{X.~Zhang}, \bibinfo{author}{X.~Huang}, \bibinfo{author}{X.~Dong}, \bibinfo{author}{Y.~Xu}, \bibinfo{author}{Y.~Wei}, \bibinfo{author}{Y.~An}, \bibinfo{author}{Y.~Niu}, \bibinfo{author}{Y.~Zhu}, \bibinfo{author}{Y.~Wen}, \bibinfo{author}{Y.~Cen}, \bibinfo{author}{Y.~Bai}, \bibinfo{author}{Z.~Qiao}, \bibinfo{author}{Z.~Wang}, \bibinfo{author}{Z.~Wang}, \bibinfo{author}{Z.~Zhu}, \bibinfo{author}{Z.~Liu}, \bibinfo{author}{Z.~Li}, \bibinfo{author}{B.~Wang}, \bibinfo{author}{B.~Wen}, \bibinfo{author}{C.~Huang}, \bibinfo{author}{C.~Cai}, \bibinfo{author}{C.~Yu}, \bibinfo{author}{C.~Li}, \bibinfo{author}{C.~Hu}, \bibinfo{author}{C.~Zhang}, \bibinfo{author}{D.~Zhang}, \bibinfo{author}{D.~Lin}, \bibinfo{author}{D.~Yang}, \bibinfo{author}{D.~Wang}, \bibinfo{author}{D.~Ai}, \bibinfo{author}{E.~Zhu}, \bibinfo{author}{F.~Yi}, \bibinfo{author}{F.~Chen}, \bibinfo{author}{G.~Wen}, \bibinfo{author}{H.~Sun}, \bibinfo{author}{H.~Zhao},
  \bibinfo{author}{H.~Hu}, \bibinfo{author}{H.~Zhang}, \bibinfo{author}{H.~Liu}, \bibinfo{author}{H.~Zhang}, \bibinfo{author}{H.~Peng}, \bibinfo{author}{H.~Tai}, \bibinfo{author}{H.~Zhang}, \bibinfo{author}{H.~Liu}, \bibinfo{author}{H.~Wang}, \bibinfo{author}{H.~Yan}, \bibinfo{author}{H.~Ge}, \bibinfo{author}{H.~Liu}, \bibinfo{author}{H.~Chu}, \bibinfo{author}{J.~Zhao}, \bibinfo{author}{J.~Wang}, \bibinfo{author}{J.~Zhao}, \bibinfo{author}{J.~Ren}, \bibinfo{author}{J.~Wang}, \bibinfo{author}{J.~Zhang}, \bibinfo{author}{J.~Gui}, \bibinfo{author}{J.~Zhao}, \bibinfo{author}{J.~Li}, \bibinfo{author}{J.~An}, \bibinfo{author}{J.~Li}, \bibinfo{author}{J.~Yuan}, \bibinfo{author}{J.~Du}, \bibinfo{author}{J.~Liu}, \bibinfo{author}{J.~Zhi}, \bibinfo{author}{J.~Duan}, \bibinfo{author}{K.~Zhou}, \bibinfo{author}{K.~Wei}, \bibinfo{author}{K.~Wang}, \bibinfo{author}{K.~Luo}, \bibinfo{author}{L.~Zhang}, \bibinfo{author}{L.~Sha}, \bibinfo{author}{L.~Xu}, \bibinfo{author}{L.~Wu}, \bibinfo{author}{L.~Ding},
  \bibinfo{author}{L.~Chen}, \bibinfo{author}{M.~Li}, \bibinfo{author}{N.~Lin}, \bibinfo{author}{P.~Ta}, \bibinfo{author}{Q.~Zou}, \bibinfo{author}{R.~Song}, \bibinfo{author}{R.~Yang}, \bibinfo{author}{S.~Tu}, \bibinfo{author}{S.~Yang}, \bibinfo{author}{S.~Wu}, \bibinfo{author}{S.~Zhang}, \bibinfo{author}{S.~Li}, \bibinfo{author}{S.~Li}, \bibinfo{author}{S.~Fan}, \bibinfo{author}{W.~Qin}, \bibinfo{author}{W.~Tian}, \bibinfo{author}{W.~Zhang}, \bibinfo{author}{W.~Yu}, \bibinfo{author}{W.~Liang}, \bibinfo{author}{X.~Kuang}, \bibinfo{author}{X.~Cheng}, \bibinfo{author}{X.~Li}, \bibinfo{author}{X.~Yan}, \bibinfo{author}{X.~Hu}, \bibinfo{author}{X.~Ling}, \bibinfo{author}{X.~Fan}, \bibinfo{author}{X.~Xia}, \bibinfo{author}{X.~Zhang}, \bibinfo{author}{X.~Zhang}, \bibinfo{author}{X.~Pan}, \bibinfo{author}{X.~Zou}, \bibinfo{author}{X.~Zhang}, \bibinfo{author}{Y.~Liu}, \bibinfo{author}{Y.~Wu}, \bibinfo{author}{Y.~Li}, \bibinfo{author}{Y.~Wang}, \bibinfo{author}{Y.~Zhu}, \bibinfo{author}{Y.~Tan},
  \bibinfo{author}{Y.~Zhou}, \bibinfo{author}{Y.~Pan}, \bibinfo{author}{Y.~Zhang}, \bibinfo{author}{Y.~Su}, \bibinfo{author}{Y.~Geng}, \bibinfo{author}{Y.~Yan}, \bibinfo{author}{Y.~Tan}, \bibinfo{author}{Y.~Bi}, \bibinfo{author}{Y.~Shen}, \bibinfo{author}{Y.~Yang}, \bibinfo{author}{Y.~Li}, \bibinfo{author}{Y.~Liu}, \bibinfo{author}{Y.~Wang}, \bibinfo{author}{Y.~Li}, \bibinfo{author}{Y.~Wu}, \bibinfo{author}{Y.~Zhang}, \bibinfo{author}{Y.~Duan}, \bibinfo{author}{Y.~Zhang}, \bibinfo{author}{Z.~Liu}, \bibinfo{author}{Z.~Jiang}, \bibinfo{author}{Z.~Yan}, \bibinfo{author}{Z.~Zhang}, \bibinfo{author}{Z.~Wei}, \bibinfo{author}{Z.~Chen}, \bibinfo{author}{Z.~Feng}, \bibinfo{author}{Z.~Yao}, \bibinfo{author}{Z.~Chai}, \bibinfo{author}{Z.~Wang}, \bibinfo{author}{Z.~Zhang}, \bibinfo{author}{B.~Xu}, \bibinfo{author}{M.~Huang}, \bibinfo{author}{H.~Wang}, \bibinfo{author}{J.~Li}, \bibinfo{author}{Y.~Dong}, \bibinfo{author}{J.~Tang}, \bibinfo{title}{Glm-5: from vibe coding to agentic engineering}, \bibinfo{year}{2026}.
  \URLprefix \url{https://arxiv.org/abs/2602.15763}. \href{http://arxiv.org/abs/2602.15763}{{\tt arXiv:2602.15763}}.
\bibitem[{MiniMax et~al.(2026)MiniMax, :, Chen, Li, Zhou, Gong, Jiang, Dan, Yu, Wang, Ma, Zhong, Zhu, Xiao, Yang, Du, Zhang, Zhang, Huang, Zhang, Du, Zhao, Guo, Chen, Ding, Sun, Zhang, Yang, Yu, Zheng, Zheng, Li, Zhu, Zhou, Zhang, Ding, Zhang, Sun, Lyu, Lu, Wang, Shi, Li, Chen, Zhang, Zhuang, Cai, Pan, Li, Song, Zhang, Wang, Gu, Zhu, Dong, Li, Zhang, Zhuang, Tian, Liu, Hu, Tao, Zhang, Ruan, Xu, Yan, Liu, He, Xu, Ji, Yang, Xiao, Duan, Li, Han, Ruan, Yuan, Yu, Feng, Mo, Li, Bao, Yang, Zhou, Loki, Chen, Ceng, Li, Zhong, Tao, Chi, Lin, Hu, Chen, Zhu, Gao, Gao, Li, Li, Zhao, Ren, Xu, Ren, Li, Wang, Chen, Ceng, Tian, Dong, Leng, Zhang, Liu, Chen, Jia, Yao, Zhao, Yu, Li, Pan, Zhu, Li, Xie, Qin, Liang, Liu, Xu, Li, Chen, Cheng, Zhang, Chen, Zhao, Chen, Song, Wang, Luo, Su, Li, Han, Wu, Song, Han, Guan, Lu, Zou, Lai, Li, Gong, Wang, Xu, Wang, Tang, Chen, Qiu, Shi, Guo, Huang, Wang, Hu, Gao, Zhang, Ying, Zhang, Wang, Song, Yang, Meng, Miao, Li, Liu, Hu, Huang, Li, Huang, Zhang, Hong, Xie, Zhang, Liao, Shi, Wenren, Li,
  Li, Luo, Jin, Sun, Zhou, Su, Li, Zhu, Peng, Fan, Zhang, Xu, Lv, Xu, He, He, Li, Gao, Wu, Song, Zhou, Sun, Huang, Chen, and Ge}]{minimax2026minimaxm2seriesminiactivations}
\bibinfo{author}{MiniMax}, \bibinfo{author}{:}, \bibinfo{author}{A.~Chen}, \bibinfo{author}{A.~Li}, \bibinfo{author}{B.~Zhou}, \bibinfo{author}{B.~Gong}, \bibinfo{author}{B.~Jiang}, \bibinfo{author}{B.~Dan}, \bibinfo{author}{C.~Yu}, \bibinfo{author}{C.~Wang}, \bibinfo{author}{C.~Ma}, \bibinfo{author}{C.~Zhong}, \bibinfo{author}{C.~Zhu}, \bibinfo{author}{C.~Xiao}, \bibinfo{author}{C.~Yang}, \bibinfo{author}{C.~Du}, \bibinfo{author}{C.~Zhang}, \bibinfo{author}{C.~Zhang}, \bibinfo{author}{C.~Huang}, \bibinfo{author}{C.~Zhang}, \bibinfo{author}{C.~Du}, \bibinfo{author}{C.~Zhao}, \bibinfo{author}{C.~Guo}, \bibinfo{author}{D.~Chen}, \bibinfo{author}{D.~Ding}, \bibinfo{author}{D.~Sun}, \bibinfo{author}{D.~Zhang}, \bibinfo{author}{E.~Yang}, \bibinfo{author}{F.~Yu}, \bibinfo{author}{G.~Zheng}, \bibinfo{author}{G.~Zheng}, \bibinfo{author}{G.~Li}, \bibinfo{author}{H.~Zhu}, \bibinfo{author}{H.~Zhou}, \bibinfo{author}{H.~Zhang}, \bibinfo{author}{H.~Ding}, \bibinfo{author}{H.~Zhang}, \bibinfo{author}{H.~Sun},
  \bibinfo{author}{H.~Lyu}, \bibinfo{author}{H.~Lu}, \bibinfo{author}{H.~Wang}, \bibinfo{author}{H.~Shi}, \bibinfo{author}{H.~Li}, \bibinfo{author}{J.~Chen}, \bibinfo{author}{J.~Zhang}, \bibinfo{author}{J.~Zhuang}, \bibinfo{author}{J.~Cai}, \bibinfo{author}{J.~Pan}, \bibinfo{author}{J.~Li}, \bibinfo{author}{J.~Song}, \bibinfo{author}{J.~Zhang}, \bibinfo{author}{J.~Wang}, \bibinfo{author}{J.~Gu}, \bibinfo{author}{J.~Zhu}, \bibinfo{author}{J.~Dong}, \bibinfo{author}{J.~Li}, \bibinfo{author}{J.~Zhang}, \bibinfo{author}{J.~Zhuang}, \bibinfo{author}{J.~Tian}, \bibinfo{author}{J.~Liu}, \bibinfo{author}{J.~Hu}, \bibinfo{author}{J.~Tao}, \bibinfo{author}{J.~Zhang}, \bibinfo{author}{J.~Ruan}, \bibinfo{author}{J.~Xu}, \bibinfo{author}{J.~Yan}, \bibinfo{author}{J.~Liu}, \bibinfo{author}{J.~He}, \bibinfo{author}{K.~Xu}, \bibinfo{author}{K.~Ji}, \bibinfo{author}{K.~Yang}, \bibinfo{author}{K.~Xiao}, \bibinfo{author}{K.~Duan}, \bibinfo{author}{K.~Li}, \bibinfo{author}{L.~Han}, \bibinfo{author}{L.~Ruan},
  \bibinfo{author}{L.~Yuan}, \bibinfo{author}{L.~Yu}, \bibinfo{author}{L.~Feng}, \bibinfo{author}{L.~Mo}, \bibinfo{author}{L.~Li}, \bibinfo{author}{L.~Bao}, \bibinfo{author}{L.~Yang}, \bibinfo{author}{L.~Zhou}, \bibinfo{author}{Loki}, \bibinfo{author}{L.~Chen}, \bibinfo{author}{L.~Ceng}, \bibinfo{author}{M.~Li}, \bibinfo{author}{M.~Zhong}, \bibinfo{author}{M.~Tao}, \bibinfo{author}{M.~Chi}, \bibinfo{author}{M.~Lin}, \bibinfo{author}{N.~Hu}, \bibinfo{author}{N.~Chen}, \bibinfo{author}{P.~Zhu}, \bibinfo{author}{P.~Gao}, \bibinfo{author}{P.~Gao}, \bibinfo{author}{P.~Li}, \bibinfo{author}{P.~Li}, \bibinfo{author}{P.~Zhao}, \bibinfo{author}{Q.~Ren}, \bibinfo{author}{Q.~Xu}, \bibinfo{author}{Q.~Ren}, \bibinfo{author}{Q.~Li}, \bibinfo{author}{Q.~Wang}, \bibinfo{author}{Q.~Chen}, \bibinfo{author}{Q.~Ceng}, \bibinfo{author}{R.~Tian}, \bibinfo{author}{R.~Dong}, \bibinfo{author}{R.~Leng}, \bibinfo{author}{R.~Zhang}, \bibinfo{author}{S.~Liu}, \bibinfo{author}{S.~Chen}, \bibinfo{author}{S.~Jia}, \bibinfo{author}{S.~Yao},
  \bibinfo{author}{S.~Zhao}, \bibinfo{author}{S.~Yu}, \bibinfo{author}{S.~Li}, \bibinfo{author}{S.~Pan}, \bibinfo{author}{S.~Zhu}, \bibinfo{author}{T.~Li}, \bibinfo{author}{T.~Xie}, \bibinfo{author}{T.~Qin}, \bibinfo{author}{T.~Liang}, \bibinfo{author}{W.~Liu}, \bibinfo{author}{W.~Xu}, \bibinfo{author}{W.~Li}, \bibinfo{author}{W.~Chen}, \bibinfo{author}{W.~Cheng}, \bibinfo{author}{W.~Zhang}, \bibinfo{author}{W.~Chen}, \bibinfo{author}{W.~Zhao}, \bibinfo{author}{X.~Chen}, \bibinfo{author}{X.~Song}, \bibinfo{author}{X.~Wang}, \bibinfo{author}{X.~Luo}, \bibinfo{author}{X.~Su}, \bibinfo{author}{X.~Li}, \bibinfo{author}{X.~Han}, \bibinfo{author}{X.~Wu}, \bibinfo{author}{X.~Song}, \bibinfo{author}{X.~Han}, \bibinfo{author}{X.~Guan}, \bibinfo{author}{X.~Lu}, \bibinfo{author}{X.~Zou}, \bibinfo{author}{X.~Lai}, \bibinfo{author}{X.~Li}, \bibinfo{author}{Y.~Gong}, \bibinfo{author}{Y.~Wang}, \bibinfo{author}{Y.~Xu}, \bibinfo{author}{Y.~Wang}, \bibinfo{author}{Y.~Tang}, \bibinfo{author}{Y.~Chen}, \bibinfo{author}{Y.~Qiu},
  \bibinfo{author}{Y.~Shi}, \bibinfo{author}{Y.~Guo}, \bibinfo{author}{Y.~Huang}, \bibinfo{author}{Y.~Wang}, \bibinfo{author}{Y.~Hu}, \bibinfo{author}{Y.~Gao}, \bibinfo{author}{Y.~Zhang}, \bibinfo{author}{Y.~Ying}, \bibinfo{author}{Y.~Zhang}, \bibinfo{author}{Y.~Wang}, \bibinfo{author}{Y.~Song}, \bibinfo{author}{Y.~Yang}, \bibinfo{author}{Y.~Meng}, \bibinfo{author}{Y.~Miao}, \bibinfo{author}{Y.~Li}, \bibinfo{author}{Y.~Liu}, \bibinfo{author}{Y.~Hu}, \bibinfo{author}{Y.~Huang}, \bibinfo{author}{Y.~Li}, \bibinfo{author}{Y.~Huang}, \bibinfo{author}{Y.~Zhang}, \bibinfo{author}{Y.~Hong}, \bibinfo{author}{Y.~Xie}, \bibinfo{author}{Y.~Zhang}, \bibinfo{author}{Y.~Liao}, \bibinfo{author}{Y.~Shi}, \bibinfo{author}{Y.~Wenren}, \bibinfo{author}{Z.~Li}, \bibinfo{author}{Z.~Li}, \bibinfo{author}{Z.~Luo}, \bibinfo{author}{Z.~Jin}, \bibinfo{author}{Z.~Sun}, \bibinfo{author}{Z.~Zhou}, \bibinfo{author}{Z.~Su}, \bibinfo{author}{Z.~Li}, \bibinfo{author}{Z.~Zhu}, \bibinfo{author}{Z.~Peng}, \bibinfo{author}{Z.~Fan},
  \bibinfo{author}{Z.~Zhang}, \bibinfo{author}{Z.~Xu}, \bibinfo{author}{Z.~Lv}, \bibinfo{author}{Z.~Xu}, \bibinfo{author}{Z.~He}, \bibinfo{author}{Z.~He}, \bibinfo{author}{Z.~Li}, \bibinfo{author}{Z.~Gao}, \bibinfo{author}{Z.~Wu}, \bibinfo{author}{Z.~Song}, \bibinfo{author}{Z.~Zhou}, \bibinfo{author}{Z.~Sun}, \bibinfo{author}{Z.~Huang}, \bibinfo{author}{Z.~Chen}, \bibinfo{author}{Z.~Ge}, \bibinfo{title}{The minimax-m2 series: Mini activations unleashing max real-world intelligence}, \bibinfo{year}{2026}. \URLprefix \url{https://arxiv.org/abs/2605.26494}. \href{http://arxiv.org/abs/2605.26494}{{\tt arXiv:2605.26494}}.
\bibitem[{Labs et~al.(2025)Labs, Khanna, Kharbanda, Li, Varma, Wang, Birnbaum, Luo, Miraoui, Palrecha, Ermon, Grover, and Kuleshov}]{labs2025mercuryultrafastlanguagemodels}
\bibinfo{author}{I.~Labs}, \bibinfo{author}{S.~Khanna}, \bibinfo{author}{S.~Kharbanda}, \bibinfo{author}{S.~Li}, \bibinfo{author}{H.~Varma}, \bibinfo{author}{E.~Wang}, \bibinfo{author}{S.~Birnbaum}, \bibinfo{author}{Z.~Luo}, \bibinfo{author}{Y.~Miraoui}, \bibinfo{author}{A.~Palrecha}, \bibinfo{author}{S.~Ermon}, \bibinfo{author}{A.~Grover}, \bibinfo{author}{V.~Kuleshov}, \bibinfo{title}{Mercury: Ultra-fast language models based on diffusion}, \bibinfo{year}{2025}. \URLprefix \url{https://arxiv.org/abs/2506.17298}. \href{http://arxiv.org/abs/2506.17298}{{\tt arXiv:2506.17298}}.
\bibitem[{OpenAI(2026)}]{gpt54nano}
\bibinfo{author}{OpenAI}, \bibinfo{title}{{I}ntroducing {G}{P}{T}-5.4 mini and nano --- openai.com}, \bibinfo{howpublished}{\url{https://openai.com/index/introducing-gpt-5-4-mini-and-nano/}}, \bibinfo{year}{2026}. \bibinfo{note}{[Accessed 28-05-2026]}.
\bibitem[{Team et~al.(2026)Team, Bai, Bai, Bao, Cai, Cao, Charles, Che, Chen, Chen, Chen, Chen, Chen, Chen, Chen, Chen, Chen, Chen, Chen, Chen, Chen, Chen, Chen, Chen, Chen, Chen, Chen, Chen, Chen, Cheng, Chu, Cui, Deng, Diao, Ding, Dong, Dong, Dong, Dong, Du, Du, Du, Du, Du, Fan, Fang, Feng, Feng, Fu, Fu, Gao, Gao, Ge, Geng, Gong, Gong, Gongque, Gu, Gu, Gu, Guan, Guo, Hao, He, He, He, Hong, Hu, Hu, Hu, Hu, Huang, Huang, Huang, Huang, Jiang, Jiang, Jin, Jing, Lai, Li, Li, Li, Li, Li, Li, Li, Li, Li, Li, Li, Li, Li, Li, Li, Li, Li, Li, Li, Li, Li, Li, Li, Liao, Lin, Lin, Lin, Lin, Liu, Liu, Liu, Liu, Liu, Liu, Liu, Liu, Liu, Liu, Liu, Liu, Liu, Liu, Liu, Liu, Liu, Liu, Lu, Lu, Lu, Luo, Luo, Luo, Ma, Ma, Mao, Mei, Men, Meng, Meng, Miao, Ni, Ouyang, Pan, Pang, Qian, Qin, Qin, Qiu, Qu, Shang, Shao, Shen, Shen, Shi, Shi, Shi, Song, Song, Song, Song, Su, Su, Su, Sui, Sun, Sun, Sun, Sung, Tai, Tang, Tang, Tang, Tang, Tao, Teng, Tian, Tian, Wang, Wang, Wang, Wang, Wang, Wang, Wang, Wang, Wang, Wang, Wang, Wang, Wang,
  Wang, Wang, Wang, Wang, Wang, Wang, Wang, Wang, Wang, Wang, Wang, Wang, Wang, Wang, Wang, Wang, Wang, Wang, Wang, Wang, Wang, Wang, Wang, Wang, Wang, Wei, Wei, Wen, Wen, Wu, Wu, Wu, Wu, Wu, Wu, Wu, Wu, Wu, Xiao, Xie, Xie, Xie, Xin, Xing, Xu, Xu, Xu, Xu, Xu, Xu, Xu, Xu, Xu, Xu, Xu, Xu, Xu, Xu, Xu, Yan, Yan, Yang, Yang, Yang, Yang, Yang, Yang, Yang, Yang, Yang, Yang, Yang, Yang, Yang, Yang, Yao, Ye, Ye, Ye, Yin, Yu, Yu, Yu, Yu, Yuan, Yuan, Yuan, Yue, Zeng, Zha, Zhan, Zhang, Zhang, Zhang, Zhang, Zhang, Zhang, Zhang, Zhang, Zhang, Zhang, Zhang, Zhang, Zhang, Zhang, Zhang, Zhang, Zhang, Zhang, Zhao, Zhao, Zhao, Zhao, Zhao, Zhao, Zhao, Zheng, Zheng, Zheng, Zheng, Zhong, Zhong, Zhong, Zhou, Zhou, Zhou, Zhou, Zhu, Zhu, Zhu, Zhu, Zhu, Zhuang, Zhuang, Zou, and Zu}]{kimiteam2026kimik25visualagentic}
\bibinfo{author}{K.~Team}, \bibinfo{author}{T.~Bai}, \bibinfo{author}{Y.~Bai}, \bibinfo{author}{Y.~Bao}, \bibinfo{author}{S.~H. Cai}, \bibinfo{author}{Y.~Cao}, \bibinfo{author}{Y.~Charles}, \bibinfo{author}{H.~S. Che}, \bibinfo{author}{C.~Chen}, \bibinfo{author}{G.~Chen}, \bibinfo{author}{H.~Chen}, \bibinfo{author}{J.~Chen}, \bibinfo{author}{J.~Chen}, \bibinfo{author}{J.~Chen}, \bibinfo{author}{J.~Chen}, \bibinfo{author}{K.~Chen}, \bibinfo{author}{L.~Chen}, \bibinfo{author}{R.~Chen}, \bibinfo{author}{X.~Chen}, \bibinfo{author}{Y.~Chen}, \bibinfo{author}{Y.~Chen}, \bibinfo{author}{Y.~Chen}, \bibinfo{author}{Y.~Chen}, \bibinfo{author}{Y.~Chen}, \bibinfo{author}{Y.~Chen}, \bibinfo{author}{Y.~Chen}, \bibinfo{author}{Y.~Chen}, \bibinfo{author}{Z.~Chen}, \bibinfo{author}{Z.~Chen}, \bibinfo{author}{D.~Cheng}, \bibinfo{author}{M.~Chu}, \bibinfo{author}{J.~Cui}, \bibinfo{author}{J.~Deng}, \bibinfo{author}{M.~Diao}, \bibinfo{author}{H.~Ding}, \bibinfo{author}{M.~Dong}, \bibinfo{author}{M.~Dong},
  \bibinfo{author}{Y.~Dong}, \bibinfo{author}{Y.~Dong}, \bibinfo{author}{A.~Du}, \bibinfo{author}{C.~Du}, \bibinfo{author}{D.~Du}, \bibinfo{author}{L.~Du}, \bibinfo{author}{Y.~Du}, \bibinfo{author}{Y.~Fan}, \bibinfo{author}{S.~Fang}, \bibinfo{author}{Q.~Feng}, \bibinfo{author}{Y.~Feng}, \bibinfo{author}{G.~Fu}, \bibinfo{author}{K.~Fu}, \bibinfo{author}{H.~Gao}, \bibinfo{author}{T.~Gao}, \bibinfo{author}{Y.~Ge}, \bibinfo{author}{S.~Geng}, \bibinfo{author}{C.~Gong}, \bibinfo{author}{X.~Gong}, \bibinfo{author}{Z.~Gongque}, \bibinfo{author}{Q.~Gu}, \bibinfo{author}{X.~Gu}, \bibinfo{author}{Y.~Gu}, \bibinfo{author}{L.~Guan}, \bibinfo{author}{Y.~Guo}, \bibinfo{author}{X.~Hao}, \bibinfo{author}{W.~He}, \bibinfo{author}{W.~He}, \bibinfo{author}{Y.~He}, \bibinfo{author}{C.~Hong}, \bibinfo{author}{H.~Hu}, \bibinfo{author}{J.~Hu}, \bibinfo{author}{Y.~Hu}, \bibinfo{author}{Z.~Hu}, \bibinfo{author}{K.~Huang}, \bibinfo{author}{R.~Huang}, \bibinfo{author}{W.~Huang}, \bibinfo{author}{Z.~Huang}, \bibinfo{author}{T.~Jiang},
  \bibinfo{author}{Z.~Jiang}, \bibinfo{author}{X.~Jin}, \bibinfo{author}{Y.~Jing}, \bibinfo{author}{G.~Lai}, \bibinfo{author}{A.~Li}, \bibinfo{author}{C.~Li}, \bibinfo{author}{C.~Li}, \bibinfo{author}{F.~Li}, \bibinfo{author}{G.~Li}, \bibinfo{author}{G.~Li}, \bibinfo{author}{H.~Li}, \bibinfo{author}{H.~Li}, \bibinfo{author}{J.~Li}, \bibinfo{author}{J.~Li}, \bibinfo{author}{J.~Li}, \bibinfo{author}{L.~Li}, \bibinfo{author}{M.~Li}, \bibinfo{author}{W.~Li}, \bibinfo{author}{W.~Li}, \bibinfo{author}{X.~Li}, \bibinfo{author}{X.~Li}, \bibinfo{author}{Y.~Li}, \bibinfo{author}{Y.~Li}, \bibinfo{author}{Y.~Li}, \bibinfo{author}{Y.~Li}, \bibinfo{author}{Z.~Li}, \bibinfo{author}{Z.~Li}, \bibinfo{author}{W.~Liao}, \bibinfo{author}{J.~Lin}, \bibinfo{author}{X.~Lin}, \bibinfo{author}{Z.~Lin}, \bibinfo{author}{Z.~Lin}, \bibinfo{author}{C.~Liu}, \bibinfo{author}{C.~Liu}, \bibinfo{author}{H.~Liu}, \bibinfo{author}{L.~Liu}, \bibinfo{author}{S.~Liu}, \bibinfo{author}{S.~Liu}, \bibinfo{author}{S.~Liu}, \bibinfo{author}{T.~Liu},
  \bibinfo{author}{T.~Liu}, \bibinfo{author}{W.~Liu}, \bibinfo{author}{X.~Liu}, \bibinfo{author}{Y.~Liu}, \bibinfo{author}{Y.~Liu}, \bibinfo{author}{Y.~Liu}, \bibinfo{author}{Y.~Liu}, \bibinfo{author}{Y.~Liu}, \bibinfo{author}{Z.~Liu}, \bibinfo{author}{Z.~Liu}, \bibinfo{author}{E.~Lu}, \bibinfo{author}{H.~Lu}, \bibinfo{author}{Z.~Lu}, \bibinfo{author}{J.~Luo}, \bibinfo{author}{T.~Luo}, \bibinfo{author}{Y.~Luo}, \bibinfo{author}{L.~Ma}, \bibinfo{author}{Y.~Ma}, \bibinfo{author}{S.~Mao}, \bibinfo{author}{Y.~Mei}, \bibinfo{author}{X.~Men}, \bibinfo{author}{F.~Meng}, \bibinfo{author}{Z.~Meng}, \bibinfo{author}{Y.~Miao}, \bibinfo{author}{M.~Ni}, \bibinfo{author}{K.~Ouyang}, \bibinfo{author}{S.~Pan}, \bibinfo{author}{B.~Pang}, \bibinfo{author}{Y.~Qian}, \bibinfo{author}{R.~Qin}, \bibinfo{author}{Z.~Qin}, \bibinfo{author}{J.~Qiu}, \bibinfo{author}{B.~Qu}, \bibinfo{author}{Z.~Shang}, \bibinfo{author}{Y.~Shao}, \bibinfo{author}{T.~Shen}, \bibinfo{author}{Z.~Shen}, \bibinfo{author}{J.~Shi}, \bibinfo{author}{L.~Shi},
  \bibinfo{author}{S.~Shi}, \bibinfo{author}{F.~Song}, \bibinfo{author}{P.~Song}, \bibinfo{author}{T.~Song}, \bibinfo{author}{X.~Song}, \bibinfo{author}{H.~Su}, \bibinfo{author}{J.~Su}, \bibinfo{author}{Z.~Su}, \bibinfo{author}{L.~Sui}, \bibinfo{author}{J.~Sun}, \bibinfo{author}{J.~Sun}, \bibinfo{author}{T.~Sun}, \bibinfo{author}{F.~Sung}, \bibinfo{author}{Y.~Tai}, \bibinfo{author}{C.~Tang}, \bibinfo{author}{H.~Tang}, \bibinfo{author}{X.~Tang}, \bibinfo{author}{Z.~Tang}, \bibinfo{author}{J.~Tao}, \bibinfo{author}{S.~Teng}, \bibinfo{author}{C.~Tian}, \bibinfo{author}{P.~Tian}, \bibinfo{author}{A.~Wang}, \bibinfo{author}{B.~Wang}, \bibinfo{author}{C.~Wang}, \bibinfo{author}{C.~Wang}, \bibinfo{author}{C.~Wang}, \bibinfo{author}{D.~Wang}, \bibinfo{author}{D.~Wang}, \bibinfo{author}{D.~Wang}, \bibinfo{author}{F.~Wang}, \bibinfo{author}{H.~Wang}, \bibinfo{author}{H.~Wang}, \bibinfo{author}{H.~Wang}, \bibinfo{author}{H.~Wang}, \bibinfo{author}{H.~Wang}, \bibinfo{author}{J.~Wang}, \bibinfo{author}{J.~Wang},
  \bibinfo{author}{J.~Wang}, \bibinfo{author}{K.~Wang}, \bibinfo{author}{L.~Wang}, \bibinfo{author}{Q.~Wang}, \bibinfo{author}{S.~Wang}, \bibinfo{author}{S.~Wang}, \bibinfo{author}{S.~Wang}, \bibinfo{author}{W.~Wang}, \bibinfo{author}{X.~Wang}, \bibinfo{author}{X.~Wang}, \bibinfo{author}{Y.~Wang}, \bibinfo{author}{Y.~Wang}, \bibinfo{author}{Y.~Wang}, \bibinfo{author}{Y.~Wang}, \bibinfo{author}{Y.~Wang}, \bibinfo{author}{Y.~Wang}, \bibinfo{author}{Z.~Wang}, \bibinfo{author}{Z.~Wang}, \bibinfo{author}{Z.~Wang}, \bibinfo{author}{Z.~Wang}, \bibinfo{author}{Z.~Wang}, \bibinfo{author}{Z.~Wang}, \bibinfo{author}{C.~Wei}, \bibinfo{author}{M.~Wei}, \bibinfo{author}{C.~Wen}, \bibinfo{author}{Z.~Wen}, \bibinfo{author}{C.~Wu}, \bibinfo{author}{H.~Wu}, \bibinfo{author}{J.~Wu}, \bibinfo{author}{R.~Wu}, \bibinfo{author}{W.~Wu}, \bibinfo{author}{Y.~Wu}, \bibinfo{author}{Y.~Wu}, \bibinfo{author}{Y.~Wu}, \bibinfo{author}{Z.~Wu}, \bibinfo{author}{C.~Xiao}, \bibinfo{author}{J.~Xie}, \bibinfo{author}{X.~Xie},
  \bibinfo{author}{Y.~Xie}, \bibinfo{author}{Y.~Xin}, \bibinfo{author}{B.~Xing}, \bibinfo{author}{B.~Xu}, \bibinfo{author}{J.~Xu}, \bibinfo{author}{J.~Xu}, \bibinfo{author}{J.~Xu}, \bibinfo{author}{L.~H. Xu}, \bibinfo{author}{L.~Xu}, \bibinfo{author}{S.~Xu}, \bibinfo{author}{W.~Xu}, \bibinfo{author}{X.~Xu}, \bibinfo{author}{X.~Xu}, \bibinfo{author}{Y.~Xu}, \bibinfo{author}{Y.~Xu}, \bibinfo{author}{Y.~Xu}, \bibinfo{author}{Z.~Xu}, \bibinfo{author}{Z.~Xu}, \bibinfo{author}{J.~Yan}, \bibinfo{author}{Y.~Yan}, \bibinfo{author}{G.~Yang}, \bibinfo{author}{H.~Yang}, \bibinfo{author}{J.~Yang}, \bibinfo{author}{K.~Yang}, \bibinfo{author}{N.~Yang}, \bibinfo{author}{R.~Yang}, \bibinfo{author}{X.~Yang}, \bibinfo{author}{X.~Yang}, \bibinfo{author}{Y.~Yang}, \bibinfo{author}{Y.~Yang}, \bibinfo{author}{Y.~Yang}, \bibinfo{author}{Z.~Yang}, \bibinfo{author}{Z.~Yang}, \bibinfo{author}{Z.~Yang}, \bibinfo{author}{H.~Yao}, \bibinfo{author}{D.~Ye}, \bibinfo{author}{W.~Ye}, \bibinfo{author}{Z.~Ye}, \bibinfo{author}{B.~Yin},
  \bibinfo{author}{C.~Yu}, \bibinfo{author}{L.~Yu}, \bibinfo{author}{T.~Yu}, \bibinfo{author}{T.~Yu}, \bibinfo{author}{E.~Yuan}, \bibinfo{author}{M.~Yuan}, \bibinfo{author}{X.~Yuan}, \bibinfo{author}{Y.~Yue}, \bibinfo{author}{W.~Zeng}, \bibinfo{author}{D.~Zha}, \bibinfo{author}{H.~Zhan}, \bibinfo{author}{D.~Zhang}, \bibinfo{author}{H.~Zhang}, \bibinfo{author}{J.~Zhang}, \bibinfo{author}{P.~Zhang}, \bibinfo{author}{Q.~Zhang}, \bibinfo{author}{R.~Zhang}, \bibinfo{author}{X.~Zhang}, \bibinfo{author}{Y.~Zhang}, \bibinfo{author}{Y.~Zhang}, \bibinfo{author}{Y.~Zhang}, \bibinfo{author}{Y.~Zhang}, \bibinfo{author}{Y.~Zhang}, \bibinfo{author}{Y.~Zhang}, \bibinfo{author}{Y.~Zhang}, \bibinfo{author}{Y.~Zhang}, \bibinfo{author}{Y.~Zhang}, \bibinfo{author}{Y.~Zhang}, \bibinfo{author}{Z.~Zhang}, \bibinfo{author}{C.~Zhao}, \bibinfo{author}{F.~Zhao}, \bibinfo{author}{J.~Zhao}, \bibinfo{author}{S.~Zhao}, \bibinfo{author}{X.~Zhao}, \bibinfo{author}{Y.~Zhao}, \bibinfo{author}{Z.~Zhao}, \bibinfo{author}{H.~Zheng},
  \bibinfo{author}{R.~Zheng}, \bibinfo{author}{S.~Zheng}, \bibinfo{author}{T.~Zheng}, \bibinfo{author}{J.~Zhong}, \bibinfo{author}{L.~Zhong}, \bibinfo{author}{W.~Zhong}, \bibinfo{author}{M.~Zhou}, \bibinfo{author}{R.~Zhou}, \bibinfo{author}{X.~Zhou}, \bibinfo{author}{Z.~Zhou}, \bibinfo{author}{J.~Zhu}, \bibinfo{author}{L.~Zhu}, \bibinfo{author}{X.~Zhu}, \bibinfo{author}{Y.~Zhu}, \bibinfo{author}{Z.~Zhu}, \bibinfo{author}{J.~Zhuang}, \bibinfo{author}{W.~Zhuang}, \bibinfo{author}{Y.~Zou}, \bibinfo{author}{X.~Zu}, \bibinfo{title}{Kimi k2.5: Visual agentic intelligence}, \bibinfo{year}{2026}. \URLprefix \url{https://arxiv.org/abs/2602.02276}. \href{http://arxiv.org/abs/2602.02276}{{\tt arXiv:2602.02276}}.
\bibitem[{TogetherAI(2026)}]{togetherTogetherNative}
\bibinfo{author}{TogetherAI}, \bibinfo{title}{{T}ogether {A}{I} | {T}he {A}{I} {N}ative {C}loud --- together.ai}, \bibinfo{howpublished}{\url{https://www.together.ai/}}, \bibinfo{year}{2026}. \bibinfo{note}{[Accessed 28-05-2026]}.
\bibitem[{FireworksAI(2026)}]{fireworksFireworksFastest}
\bibinfo{author}{FireworksAI}, \bibinfo{title}{{F}ireworks {A}{I} - {F}astest {I}nference for {G}enerative {A}{I} --- fireworks.ai}, \bibinfo{howpublished}{\url{https://fireworks.ai/}}, \bibinfo{year}{2026}. \bibinfo{note}{[Accessed 28-05-2026]}.
\bibitem[{Tolstykh et~al.(2025)Tolstykh, Tsybina, Yakubson, and Kuprashevich}]{tolstykh2025llmtrace}
\bibinfo{author}{I.~Tolstykh}, \bibinfo{author}{A.~Tsybina}, \bibinfo{author}{S.~Yakubson}, \bibinfo{author}{M.~Kuprashevich},
\newblock \bibinfo{title}{Llmtrace: A corpus for classification and fine-grained localization of ai-written text},
\newblock \bibinfo{journal}{arXiv preprint arXiv:2509.21269}  (\bibinfo{year}{2025}).
\bibitem[{Gugliotta et~al.(2025)Gugliotta, La~Cava, and Tagarelli}]{gugliotta2025personagen}
\bibinfo{author}{C.~Gugliotta}, \bibinfo{author}{L.~La~Cava}, \bibinfo{author}{A.~Tagarelli},
\newblock \bibinfo{title}{Personagen: A persona-driven open-ended machine-generated text dataset},
\newblock in: \bibinfo{booktitle}{Proceedings of the 34th ACM International Conference on Information and Knowledge Management}, \bibinfo{year}{2025}, pp. \bibinfo{pages}{6397--6401}.
\bibitem[{HuggingFaceGECLM(2023)}]{huggingfaceHuggingFaceGECLMStackExchange_Mar2023Datasets}
\bibinfo{author}{HuggingFaceGECLM}, \bibinfo{title}{{H}ugging{F}ace{G}{E}{C}{L}{M}/{S}tack{E}xchange\_{M}ar2023 Â· {D}atasets at {H}ugging {F}ace --- huggingface.co}, \bibinfo{howpublished}{\url{https://huggingface.co/datasets/{H}ugging{F}ace{G}{E}{C}{L}{M}/{S}tack{E}xchange\_{M}ar2023}}, \bibinfo{year}{2023}. \bibinfo{note}{[Accessed 17-05-2026]}.
\bibitem[{Song et~al.(2026)Song, Cheng, Xu, and Feng}]{song2026maga}
\bibinfo{author}{A.~Song}, \bibinfo{author}{Y.~Cheng}, \bibinfo{author}{Y.~Xu}, \bibinfo{author}{R.~Feng},
\newblock \bibinfo{title}{Maga-bench: Machine-augment-generated text via alignment detection benchmark},
\newblock \bibinfo{journal}{arXiv preprint arXiv:2601.04633}  (\bibinfo{year}{2026}).
\bibitem[{Wu et~al.(2024)Wu, Zhan, Wong, Yang, Yang, Yuan, and Chao}]{wu2024detectrl}
\bibinfo{author}{J.~Wu}, \bibinfo{author}{R.~Zhan}, \bibinfo{author}{D.~F. Wong}, \bibinfo{author}{S.~Yang}, \bibinfo{author}{X.~Yang}, \bibinfo{author}{Y.~Yuan}, \bibinfo{author}{L.~S. Chao},
\newblock \bibinfo{title}{Detectrl: Benchmarking llm-generated text detection in real-world scenarios},
\newblock \bibinfo{journal}{Advances in Neural Information Processing Systems} \bibinfo{volume}{37} (\bibinfo{year}{2024}) \bibinfo{pages}{100369--100401}.
\bibitem[{Wang et~al.(2026)Wang, Xiong, Lian, and Dou}]{wang2026reasoningawareaigcdetectionalignment}
\bibinfo{author}{Z.~Wang}, \bibinfo{author}{M.~Xiong}, \bibinfo{author}{J.~Lian}, \bibinfo{author}{Z.~Dou}, \bibinfo{title}{Reasoning-aware aigc detection via alignment and reinforcement}, \bibinfo{year}{2026}. \URLprefix \url{https://arxiv.org/abs/2604.19172}. \href{http://arxiv.org/abs/2604.19172}{{\tt arXiv:2604.19172}}.
\bibitem[{Cava et~al.(2024)Cava, Costa, and Tagarelli}]{openturingbench}
\bibinfo{author}{L.~L. Cava}, \bibinfo{author}{D.~Costa}, \bibinfo{author}{A.~Tagarelli},
\newblock \bibinfo{title}{Is contrasting all you need? contrastive learning for the detection and attribution of ai-generated text},
\newblock in: \bibinfo{editor}{U.~Endriss}, \bibinfo{editor}{F.~S. Melo}, \bibinfo{editor}{K.~Bach}, \bibinfo{editor}{A.~J.~B. Diz}, \bibinfo{editor}{J.~M. Alonso{-}Moral}, \bibinfo{editor}{S.~Barro}, \bibinfo{editor}{F.~Heintz} (Eds.), \bibinfo{booktitle}{{ECAI} 2024 - 27th European Conference on Artificial Intelligence, 19-24 October 2024, Santiago de Compostela, Spain - Including 13th Conference on Prestigious Applications of Intelligent Systems {(PAIS} 2024)}, volume \bibinfo{volume}{392} of \textit{\bibinfo{series}{Frontiers in Artificial Intelligence and Applications}}, \bibinfo{publisher}{{IOS} Press}, \bibinfo{year}{2024}, pp. \bibinfo{pages}{3179--3186}. \URLprefix \url{https://doi.org/10.3233/FAIA240862}. \DOIprefix\doi{10.3233/FAIA240862}.
\bibitem[{Bevendorff et~al.(2025)Bevendorff, Dementieva, Fr{\"o}be, Gipp, Greiner-Petter, Karlgren, Mayerl, Nakov, Panchenko, Potthast et~al.}]{bevendorff2025overview}
\bibinfo{author}{J.~Bevendorff}, \bibinfo{author}{D.~Dementieva}, \bibinfo{author}{M.~Fr{\"o}be}, \bibinfo{author}{B.~Gipp}, \bibinfo{author}{A.~Greiner-Petter}, \bibinfo{author}{J.~Karlgren}, \bibinfo{author}{M.~Mayerl}, \bibinfo{author}{P.~Nakov}, \bibinfo{author}{A.~Panchenko}, \bibinfo{author}{M.~Potthast}, et~al.,
\newblock \bibinfo{title}{Overview of pan 2025: Generative ai detection, multilingual text detoxification, multi-author writing style analysis, and generative plagiarism detection},
\newblock in: \bibinfo{booktitle}{European Conference on Information Retrieval}, \bibinfo{organization}{Springer}, \bibinfo{year}{2025}, pp. \bibinfo{pages}{434--441}.
\bibitem[{Hamborg et~al.(2017)Hamborg, Meuschke, Breitinger, and Gipp}]{ccnews}
\bibinfo{author}{F.~Hamborg}, \bibinfo{author}{N.~Meuschke}, \bibinfo{author}{C.~Breitinger}, \bibinfo{author}{B.~Gipp},
\newblock \bibinfo{title}{news-please: A generic news crawler and extractor},
\newblock in: \bibinfo{booktitle}{Proceedings of the 15th International Symposium of Information Science}, \bibinfo{year}{2017}, pp. \bibinfo{pages}{218--223}. \DOIprefix\doi{10.5281/zenodo.4120316}.
\bibitem[{Roy et~al.(2026)Roy, Imanpour, Aziz, Bajpai, Singh, Biswas, Wanaskar, Patwa, Ghosh, Dixit, Pal, Rawte, Garimella, Jena, Sheth, Sharma, Reganti, Jain, Chadha, and Das}]{nytimesai}
\bibinfo{author}{R.~Roy}, \bibinfo{author}{N.~Imanpour}, \bibinfo{author}{A.~Aziz}, \bibinfo{author}{S.~Bajpai}, \bibinfo{author}{G.~Singh}, \bibinfo{author}{S.~Biswas}, \bibinfo{author}{K.~Wanaskar}, \bibinfo{author}{P.~Patwa}, \bibinfo{author}{S.~Ghosh}, \bibinfo{author}{S.~Dixit}, \bibinfo{author}{N.~R. Pal}, \bibinfo{author}{V.~Rawte}, \bibinfo{author}{R.~Garimella}, \bibinfo{author}{G.~Jena}, \bibinfo{author}{A.~Sheth}, \bibinfo{author}{V.~Sharma}, \bibinfo{author}{A.~N. Reganti}, \bibinfo{author}{V.~Jain}, \bibinfo{author}{A.~Chadha}, \bibinfo{author}{A.~Das}, \bibinfo{title}{A comprehensive dataset for human vs. ai generated text detection}, \bibinfo{year}{2026}. \URLprefix \url{https://arxiv.org/abs/2510.22874}. \href{http://arxiv.org/abs/2510.22874}{{\tt arXiv:2510.22874}}.
\bibitem[{Tripathi(2025)}]{sage}
\bibinfo{author}{A.~Tripathi}, \bibinfo{title}{{E}nd{L}ess{T}ime/{S}{A}{G}{E} · {D}atasets at {H}ugging {F}ace --- huggingface.co}, \bibinfo{howpublished}{\url{https://huggingface.co/datasets/{E}nd{L}ess{T}ime/{S}{A}{G}{E}}}, \bibinfo{year}{2025}. \bibinfo{note}{[Accessed 17-05-2026]}.
\bibitem[{Gromadzki et~al.(2026)Gromadzki, Wróblewska, and Kaliska}]{detectaicalibration}
\bibinfo{author}{M.~Gromadzki}, \bibinfo{author}{A.~Wróblewska}, \bibinfo{author}{A.~Kaliska}, \bibinfo{title}{On the effectiveness of llm-specific fine-tuning for detecting ai-generated text}, \bibinfo{year}{2026}. \URLprefix \url{https://arxiv.org/abs/2601.20006}. \href{http://arxiv.org/abs/2601.20006}{{\tt arXiv:2601.20006}}.
\bibitem[{Artemova et~al.(2025)Artemova, Lucas, Venkatraman, Lee, Tilga, Uchendu, and Mikhailov}]{artemova2025beemobenchmarkexperteditedmachinegenerated}
\bibinfo{author}{E.~Artemova}, \bibinfo{author}{J.~Lucas}, \bibinfo{author}{S.~Venkatraman}, \bibinfo{author}{J.~Lee}, \bibinfo{author}{S.~Tilga}, \bibinfo{author}{A.~Uchendu}, \bibinfo{author}{V.~Mikhailov}, \bibinfo{title}{Beemo: Benchmark of expert-edited machine-generated outputs}, \bibinfo{year}{2025}. \URLprefix \url{https://arxiv.org/abs/2411.04032}. \href{http://arxiv.org/abs/2411.04032}{{\tt arXiv:2411.04032}}.
\bibitem[{Chen et~al.(2025)Chen, Chen, Mo, Chen, He, Han, and Sun}]{chen2025coconutsconcentratingcontentneglecting}
\bibinfo{author}{Y.~Chen}, \bibinfo{author}{J.~Chen}, \bibinfo{author}{G.~Mo}, \bibinfo{author}{X.~Chen}, \bibinfo{author}{B.~He}, \bibinfo{author}{X.~Han}, \bibinfo{author}{L.~Sun}, \bibinfo{title}{{CoCoNUTS: Concentrating on Content while Neglecting Uninformative Textual Styles for AI-Generated Peer Review Detection}}, \bibinfo{year}{2025}. \URLprefix \url{https://arxiv.org/abs/2509.04460}. \href{http://arxiv.org/abs/2509.04460}{{\tt arXiv:2509.04460}}.
\bibitem[{Conover et~al.(2023)Conover, Hayes, Mathur, Xie, Wan, Shah, Ghodsi, Wendell, Zaharia, and Xin}]{DatabricksBlog2023DollyV2}
\bibinfo{author}{M.~Conover}, \bibinfo{author}{M.~Hayes}, \bibinfo{author}{A.~Mathur}, \bibinfo{author}{J.~Xie}, \bibinfo{author}{J.~Wan}, \bibinfo{author}{S.~Shah}, \bibinfo{author}{A.~Ghodsi}, \bibinfo{author}{P.~Wendell}, \bibinfo{author}{M.~Zaharia}, \bibinfo{author}{R.~Xin}, \bibinfo{title}{Free dolly: Introducing the world's first truly open instruction-tuned llm}, \bibinfo{year}{2023}. \URLprefix \url{https://www.databricks.com/blog/2023/04/12/dolly-first-open-commercially-viable-instruction-tuned-llm}.
\bibitem[{Ben~Allal et~al.(2024)Ben~Allal, Lozhkov, Penedo, Wolf, and von Werra}]{benallal2024cosmopedia}
\bibinfo{author}{L.~Ben~Allal}, \bibinfo{author}{A.~Lozhkov}, \bibinfo{author}{G.~Penedo}, \bibinfo{author}{T.~Wolf}, \bibinfo{author}{L.~von Werra}, \bibinfo{title}{Cosmopedia}, \bibinfo{year}{2024}. \URLprefix \url{https://huggingface.co/datasets/HuggingFaceTB/cosmopedia}.
\bibitem[{OriginalityAI(2023)}]{githubGitHubOriginalityAIAIdetectorresearchtool}
\bibinfo{author}{OriginalityAI}, \bibinfo{title}{{G}it{H}ub - {O}riginality{A}{I}/{A}{I}-detector-research-tool --- github.com}, \bibinfo{howpublished}{\url{https://github.com/{O}riginality{A}{I}/{A}{I}-detector-research-tool}}, \bibinfo{year}{2023}. \bibinfo{note}{[Accessed 18-05-2026]}.
\bibitem[{Zhu et~al.(2025)Zhu, Ren, Cao, Lin, Fang, and Li}]{zhu-etal-2025-reliably}
\bibinfo{author}{X.~Zhu}, \bibinfo{author}{Y.~Ren}, \bibinfo{author}{Y.~Cao}, \bibinfo{author}{X.~Lin}, \bibinfo{author}{F.~Fang}, \bibinfo{author}{Y.~Li},
\newblock \bibinfo{title}{Reliably bounding false positives: A zero-shot machine-generated text detection framework via multiscaled conformal prediction},
\newblock in: \bibinfo{editor}{W.~Che}, \bibinfo{editor}{J.~Nabende}, \bibinfo{editor}{E.~Shutova}, \bibinfo{editor}{M.~T. Pilehvar} (Eds.), \bibinfo{booktitle}{Proceedings of the 63rd Annual Meeting of the Association for Computational Linguistics (Volume 1: Long Papers)}, \bibinfo{publisher}{Association for Computational Linguistics}, \bibinfo{address}{Vienna, Austria}, \bibinfo{year}{2025}, pp. \bibinfo{pages}{12298--12319}. \URLprefix \url{https://aclanthology.org/2025.acl-long.601/}. \DOIprefix\doi{10.18653/v1/2025.acl-long.601}.
\bibitem[{Jabarian and Imas(2025)}]{jabarian2025artificial}
\bibinfo{author}{B.~Jabarian}, \bibinfo{author}{A.~Imas}, \bibinfo{title}{Artificial writing and automated detection}, \bibinfo{type}{Technical Report}, National Bureau of Economic Research, \bibinfo{year}{2025}.
\bibitem[{Bhargava et~al.(2021)Bhargava, Drozd, and Rogers}]{berttiny1}
\bibinfo{author}{P.~Bhargava}, \bibinfo{author}{A.~Drozd}, \bibinfo{author}{A.~Rogers}, \bibinfo{title}{Generalization in nli: Ways (not) to go beyond simple heuristics}, \bibinfo{year}{2021}. \href{http://arxiv.org/abs/2110.01518}{{\tt arXiv:2110.01518}}.
\bibitem[{Turc et~al.(2019)Turc, Chang, Lee, and Toutanova}]{berttiny2}
\bibinfo{author}{I.~Turc}, \bibinfo{author}{M.~Chang}, \bibinfo{author}{K.~Lee}, \bibinfo{author}{K.~Toutanova},
\newblock \bibinfo{title}{Well-read students learn better: The impact of student initialization on knowledge distillation},
\newblock \bibinfo{journal}{CoRR} \bibinfo{volume}{abs/1908.08962} (\bibinfo{year}{2019}). \URLprefix \url{http://arxiv.org/abs/1908.08962}. \href{http://arxiv.org/abs/1908.08962}{{\tt arXiv:1908.08962}}.
\bibitem[{Esposito(2020)}]{blitzbayesian}
\bibinfo{author}{P.~Esposito}, \bibinfo{title}{Blitz - bayesian layers in torch zoo (a bayesian deep learing library for torch)}, \bibinfo{howpublished}{\url{https://github.com/piEsposito/blitz-bayesian-deep-learning/}}, \bibinfo{year}{2020}.
\bibitem[{Blundell et~al.(2015)Blundell, Cornebise, Kavukcuoglu, and Wierstra}]{blundell2015weightuncertaintyneuralnetworks}
\bibinfo{author}{C.~Blundell}, \bibinfo{author}{J.~Cornebise}, \bibinfo{author}{K.~Kavukcuoglu}, \bibinfo{author}{D.~Wierstra}, \bibinfo{title}{Weight uncertainty in neural networks}, \bibinfo{year}{2015}. \URLprefix \url{https://arxiv.org/abs/1505.05424}. \href{http://arxiv.org/abs/1505.05424}{{\tt arXiv:1505.05424}}.
\bibitem[{Swayamdipta et~al.(2020)Swayamdipta, Schwartz, Lourie, Wang, Hajishirzi, Smith, and Choi}]{datasetcartography}
\bibinfo{author}{S.~Swayamdipta}, \bibinfo{author}{R.~Schwartz}, \bibinfo{author}{N.~Lourie}, \bibinfo{author}{Y.~Wang}, \bibinfo{author}{H.~Hajishirzi}, \bibinfo{author}{N.~A. Smith}, \bibinfo{author}{Y.~Choi},
\newblock \bibinfo{title}{Dataset cartography: Mapping and diagnosing datasets with training dynamics},
\newblock in: \bibinfo{editor}{B.~Webber}, \bibinfo{editor}{T.~Cohn}, \bibinfo{editor}{Y.~He}, \bibinfo{editor}{Y.~Liu} (Eds.), \bibinfo{booktitle}{Proceedings of the 2020 Conference on Empirical Methods in Natural Language Processing (EMNLP)}, \bibinfo{publisher}{Association for Computational Linguistics}, \bibinfo{address}{Online}, \bibinfo{year}{2020}, pp. \bibinfo{pages}{9275--9293}. \URLprefix \url{https://aclanthology.org/2020.emnlp-main.746/}. \DOIprefix\doi{10.18653/v1/2020.emnlp-main.746}.
\bibitem[{LibAUC(????)}]{libaucLibauclossesx2014}
\bibinfo{author}{LibAUC}, \bibinfo{title}{libauc.losses \&\#x2014; libauc 1.0.0 documentation --- docs.libauc.org}, \bibinfo{howpublished}{\url{https://docs.libauc.org/api/libauc.losses.html\#libauc.losses.auc.tp{A}{U}{C}\_{K}{L}\_{L}oss}}, ???? \bibinfo{note}{[Accessed 28-06-2026]}.
\bibitem[{Zhu et~al.(2022)Zhu, Li, Wang, Wu, and Yang}]{zhu2022auc}
\bibinfo{author}{D.~Zhu}, \bibinfo{author}{G.~Li}, \bibinfo{author}{B.~Wang}, \bibinfo{author}{X.~Wu}, \bibinfo{author}{T.~Yang},
\newblock \bibinfo{title}{When auc meets dro: Optimizing partial auc for deep learning with non-convex convergence guarantee},
\newblock in: \bibinfo{booktitle}{International Conference on Machine Learning}, \bibinfo{organization}{PMLR}, \bibinfo{year}{2022}, pp. \bibinfo{pages}{27548--27573}.
\bibitem[{Virtanen et~al.(2020)Virtanen, Gommers, Oliphant, Haberland, Reddy, Cournapeau, Burovski, Peterson, Weckesser, Bright, {van der Walt}, Brett, Wilson, Millman, Mayorov, Nelson, Jones, Kern, Larson, Carey, Polat, Feng, Moore, {VanderPlas}, Laxalde, Perktold, Cimrman, Henriksen, Quintero, Harris, Archibald, Ribeiro, Pedregosa, {van Mulbregt}, and {SciPy 1.0 Contributors}}]{2020SciPy-NMeth}
\bibinfo{author}{P.~Virtanen}, \bibinfo{author}{R.~Gommers}, \bibinfo{author}{T.~E. Oliphant}, \bibinfo{author}{M.~Haberland}, \bibinfo{author}{T.~Reddy}, \bibinfo{author}{D.~Cournapeau}, \bibinfo{author}{E.~Burovski}, \bibinfo{author}{P.~Peterson}, \bibinfo{author}{W.~Weckesser}, \bibinfo{author}{J.~Bright}, \bibinfo{author}{S.~J. {van der Walt}}, \bibinfo{author}{M.~Brett}, \bibinfo{author}{J.~Wilson}, \bibinfo{author}{K.~J. Millman}, \bibinfo{author}{N.~Mayorov}, \bibinfo{author}{A.~R.~J. Nelson}, \bibinfo{author}{E.~Jones}, \bibinfo{author}{R.~Kern}, \bibinfo{author}{E.~Larson}, \bibinfo{author}{C.~J. Carey}, \bibinfo{author}{{\.I}.~Polat}, \bibinfo{author}{Y.~Feng}, \bibinfo{author}{E.~W. Moore}, \bibinfo{author}{J.~{VanderPlas}}, \bibinfo{author}{D.~Laxalde}, \bibinfo{author}{J.~Perktold}, \bibinfo{author}{R.~Cimrman}, \bibinfo{author}{I.~Henriksen}, \bibinfo{author}{E.~A. Quintero}, \bibinfo{author}{C.~R. Harris}, \bibinfo{author}{A.~M. Archibald}, \bibinfo{author}{A.~H. Ribeiro},
  \bibinfo{author}{F.~Pedregosa}, \bibinfo{author}{P.~{van Mulbregt}}, \bibinfo{author}{{SciPy 1.0 Contributors}},
\newblock \bibinfo{title}{{{SciPy} 1.0: Fundamental Algorithms for Scientific Computing in Python}},
\newblock \bibinfo{journal}{Nature Methods} \bibinfo{volume}{17} (\bibinfo{year}{2020}) \bibinfo{pages}{261--272}. \DOIprefix\doi{10.1038/s41592-019-0686-2}.
\bibitem[{Lambda(2026)}]{lambdaSuperintelligenceCloud}
\bibinfo{author}{Lambda}, \bibinfo{title}{{T}he {S}uperintelligence {C}loud | {L}ambda --- lambda.ai}, \bibinfo{howpublished}{\url{https://lambda.ai/}}, \bibinfo{year}{2026}. \bibinfo{note}{[Accessed 29-05-2026]}.

\end{thebibliography}


\end{document}